# Bringing AI to Autonomous Systems -- From Cognition to Collective Intelligence

Joseph Sifakis
Verimag Laboratory, France

## Abstract

The purpose of this article is to highlight the central role of autonomous systems as the ultimate stage in the development of AI, to explain the underlying technical challenges that require a combination of connectionist AI and symbolic AI, and to integrate AI and systems engineering.

We present a comprehensive framework for the design and evaluation of autonomous systems, based on a generic agent architecture that characterizes their behavior as the composition of cognitive functions organized around a long-term memory containing the agent's evolving knowledge.

We address the challenges posed by the implementation of the fundamental features of the agent architecture, in particular the link between sensory data and structured data stored in memory, decision-making related to the achievement of the agent's goals and their planning, as well as the coordination of agents to combine individual and collective intelligence.

We explain that agent trustworthiness, unlike that of traditional systems, is not limited to behavioral properties. It includes an essential dimension related to cognitive properties, the validity of which depends on how the agent uses its knowledge in decision-making. We present avenues for the development of methods for evaluating agent trustworthiness.

We conclude with a critical assessment of the substantial gap between the aspirational vision of autonomous multi-agent systems and the current state of the art.

## 1. Introduction

Artificial intelligence is still in its infancy. Despite impressive results, culminating in the arrival of generative AI, we have yet to acquire the principles and techniques needed to synthesize intelligent systems with the same confidence and rigor with which we construct traditional systems. Today, AI provides us with the building blocks needed to create intelligent systems, but not the engineering principles required to design them reliably.

Most AI applications currently in use are assistants that interact with users through question-and-answer exchanges. These applications cover only a small fraction of the potential uses of AI in science, business, and industry. In these domains, we lack reliable solutions for building monitors that analyze data streams to generate actionable knowledge, and, most critically, autonomous systems that control complex environments and provide services with little or no human intervention.

**Autonomous Systems: The Ultimate Stage**

Autonomous systems represent the ultimate stage in AI development. They combine the characteristics of traditional automation, such as industrial robots, autopilots, and programmable logic controllers, with modern AI capabilities for processing unstructured data. Their potential applications are vast and transformative: autonomous transportation networks, resilient smart grids, self-optimizing smart factories, self-managing communication infrastructures, and smart cities [1][2].

Among all AI systems, autonomous systems most closely resemble human intelligence. They are composed of agents, each pursuing its own individual goals while coordinating to achieve the system's overall goals through interaction with a dynamic, physical, and social environment populated by other agents. This dual nature, individual intelligence and collective intelligence, is what makes them both powerful and profoundly challenging. Human societies, animal colonies, multicellular organisms, and even businesses all illustrate the paradigm of autonomy that inspires the vision of artificial autonomous systems.

**The Gap Between Vision and Reality**

Achieving this vision poses technical challenges that extend far beyond the current state of the art, which remains largely rooted in machine learning. Predictions that fully autonomous vehicles would be available by 2020 have proven to be completely wrong [3]. Despite bold claims that "2025 will be the year of AI agents," the reality is sobering: today's AI agents can provide only low-reliability services, primarily by interacting with digital worlds. They can reserve a restaurant table or summarize a document, but they fail when faced with the dynamic unpredictability of the physical world. The gap between digital capabilities and physical embodiment remains considerable.

What we need are AI systems that not only learn, but also reason; that pursue goals rationally and in accordance with technical, legal, and ethical standards; and that are sufficiently reliable to guarantee safety-critical tasks with failure rates on the order of $10^{-8}$ per hour of operation, a level of dependability currently impossible to achieve with AI components.

**The Historical Context**

The idea of creating multi-agent systems is not new. It has roots in symbolic AI research that began in the 1980s. The advent of large language models has sparked renewed commercial interest in multi-agent systems (MAS), now promoted as a central focus of AI development for autonomous systems. However, efforts to develop and implement MAS are primarily driven by marketing and strategic objectives rather than technical maturity. In their race for market dominance, major technology companies have proposed protocols such as ACP, MCP, and A2A to connect agents to tools and to one another. These protocols provide the infrastructure, the channels through which agents can communicate, but they do not solve the fundamental problems: *what* agents should communicate, *why* they would choose to do so, or *how* to ensure that their interactions are reliable.

**The Trustworthiness Challenge**

One of the main challenges in designing autonomous systems lies in the unreliability of their components. Traditional systems achieve reliability through provable properties: they behave as expected, are fault-tolerant, and have known limits. The trust we place in them is not blind; it is based on evidence. AI systems, by contrast, are opaque, empirical, and unpredictable. They learn from data rather than being defined by design. Their behavior emerges from the interaction of millions of parameters rather than explicit rules. This fundamentally changes what it means for a system to be reliable, and what it takes to demonstrate that reliability.

It is remarkable that public and private institutions are publishing guidelines, frameworks, and studies that define requirements regarding the cognitive properties of AI. However, not only do these publications make no technical contribution, but they also remain wishful thinking, impossible to implement in practice, and contribute to increasing confusion about what can reasonably be expected from AI systems and their developers [4]. Research on cognitive properties has given rise to an increasingly extensive body of literature on "responsible AI," "aligned AI," and "ethical AI" [5]. These terms feature prominently in the marketing campaigns of major technology companies, where they are presented as essential attributes of their products.

Unfortunately, current studies evaluate these properties based on behavioral criteria [6]. However, to meet cognitive properties, an agent must act intentionally, be held accountable for its actions, understand and respect the norms governing its behavior, and bear the consequences of its actions. These conditions are met by human agents, but not by current AI systems.

**Contributions of This Article**

This article addresses the gap between aspiration and reality through four main contributions:

**First**, we highlight the central role of autonomous systems as the ultimate stage in AI development, and explain the underlying technical challenges that demand a combination of connectionist AI and symbolic AI, as well as an integration of AI and systems engineering.

**Second**, we present a comprehensive framework for the design and evaluation of autonomous systems, based on an agent reference architecture that characterizes behavior as the composition of cognitive functions organized around a long-term memory containing the agent's evolving knowledge.

**Third**, we address the challenges posed by implementing the architecture's fundamental features: the link between sensory data and structured data stored in memory; decision-making related to goal achievement and planning; and the coordination of agents to combine individual and collective intelligence.

**Fourth**, we explain that agent trustworthiness, unlike that of traditional systems, is not limited to behavioral properties. It includes an essential cognitive dimension, the validity of which depends on how the agent uses its knowledge in decision-making. We present avenues for developing methods to evaluate this trustworthiness.

We conclude with a critical assessment of the substantial gap between the aspirational vision of autonomous multi-agent systems and the current state of the art.

## 2. The Agent Reference Architecture

### 2.1 The Trend Toward Architectural Solutions

For decades, the dream of autonomous systems was held back by a single, persistent obstacle: their inability to perceive. They could not recognize a pedestrian on a crowded street, understand a voice command, or decipher a handwritten note. They were blind to the chaotic, unstructured world in which humans navigate with natural ease.

Then came breakthroughs in machine learning. Suddenly, machines gained the ability to see. They could interpret images, understand speech, and process text with near-human proficiency. This was the missing piece of the puzzle: the ability to analyze complex, unstructured data. It paved the way for a goal that had long seemed out of reach, autonomous agents capable of perception and, to some extent, situational awareness.

But solving the problem of perception raises deeper questions: How does perception connect to understanding? How does it relate to a model of the world? And how does it support reasoning? An autonomous agent must not only perceive; it must also decide, plan, and act. This requires an architecture that integrates these functions. The question is: *what should this architecture look like?*

**Three Approaches to Endowing Agents with Reasoning**

There is now broad consensus that reasoning capabilities are essential for autonomous agents. However, opinions diverge sharply on how to achieve them. Three competing approaches address this question.

**Scaling up.** This approach leverages more data, computing power, and parameters to solve the problem. The idea is that reasoning will simply emerge from scale: if a sufficiently large model is trained on a sufficient volume of data, it will eventually develop common sense, the ability to reason, and perhaps even superintelligence. This technically unfounded gamble [7] is favored by the tech giants, who have the resources to play this game better than anyone else. Conveniently, it also serves as a business strategy disguised as science: it justifies multi-billion-dollar investments and creates a "compute divide" that only the wealthiest can bridge.

**Symbolic from the start.** This approach is based on the premise that symbolic reasoning must be a fundamental function from the outset. It advocates neurosymbolic AI, a hybrid approach combining neural networks with symbolic reasoning engines. Techniques include retrieval-augmented generation (RAG), which anchors large language models (LLMs) in external knowledge bases, and the integration of symbolic engines like WolframAlpha or AlphaGeometry. The core challenge lies in making these hybrids work: linking knowledge derived from data to symbolic knowledge in a seamless and reliable manner.

**Building a model of the world.** This increasingly popular approach is based on the premise that an agent cannot act intelligently without a mental model of how the world works: what changes, what remains constant, and what consequences its actions will have. A notable example is LeCun's Joint-Embedding Predictive Architecture [8], which learns abstract representations of how the world evolves, representations relevant to the target goal. Instead of predicting every pixel or every word, it predicts the essential features that matter for decision-making.

However, prediction alone is not enough. A model of the world can tell you what might happen, but it cannot tell you what you *should* do. It cannot evaluate trade-offs between competing goals, resolve conflicts between values, or exercise judgment. These are deliberative functions that require more than prediction: they require reasoning, evaluation, and choice. This is why the world-model approach, while valuable, cannot replace the rational deliberative processes required for fully autonomous agents—value-based decision-making, goal reorientation, conflict resolution, and normative reasoning—tasks that humans perform effortlessly. Prediction is necessary, but it is not sufficient.

**The Emerging Trend Toward Modular Architectures**

Despite their differences, a clear architectural trend is emerging across all three approaches. There is a gradual shift away from monolithic end-to-end solutions, which were adopted very early on by the autonomous vehicle industry, toward modular architectures. This evolution reflects the need to better understand agent behavior as a composition of elementary functions: perception, reasoning, decision-making, and both short- and long-term memory.

Proposed solutions are based on two key observations:

1. **Agents must possess knowledge** about the constantly changing external world.
2. **Agents must be able to anticipate** the behavior of their environment and plan their actions accordingly to achieve their goals.

The architecture we present is part of this modular trend. It identifies the functional components that every autonomous agent must possess, as well as the relationships that connect them through an agent knowledge base.

## 2.2 The Agent’s Knowledge Attributes: Defining the Agent's Profile

Every agent, whether a self-driving car navigating a city street, a chatbot answering a customer's question, or a police officer responding to an emergency, can be described by a few essential characteristics. These characteristics answer four fundamental questions:

- What is the agent's mission? An agent's mission constitutes its raison d'être, its reason for being, and is defined by a set of goals the agent must achieve while operating within a given environment.
- What means does the agent have at its disposal? To carry out its mission, the agent is equipped with a set of tangible and intangible resources that enable it to accomplish its mission.
- What are the conditions for the agent's harmonious integration into its environment? The agent must take into account the constraints its actions must satisfy within its physical environment, as well as the norms with which its behavior must conform in relation to its social environment.
- What are the agent's operating conditions? These are conditions related to the agent's internal state, failure to meet which could compromise its mission.

In the following subsections, we explore each of these questions in turn to provide the foundation for the agent architecture.

### *2.2.1 The Agent's Mission: What is the agent supposed to achieve?*

An agent's mission is defined with respect to the physical and social environment in which it must operate. It is characterized by a set of goals that constitute abstract specifications of the services the agent provides. The mission is the "why", the overarching purpose of the agent's existence. The goals are the "how", the specific goals that must be achieved to fulfil the mission.

For an autonomous vehicle, possible goals naturally include driving from a starting point to a destination. But this mission also includes a series of short- and medium-term goals that define the vehicle's safety and its ability to perform a wide range of maneuvers. These include, for example, avoiding collisions, keeping the vehicle on a predefined path, and performing maneuvers such as passing another car, merging onto a main road, or parking in a parking space.

The distinction between mission and goals allows for flexibility: the same mission can be achieved through different sets of goals, and goals can be added or modified without changing the mission. For example, a self-driving car's mission remains the same whether it is driving on a highway or through a city center, but the specific goals it pursues may differ.

An agent operates in an environment consisting of two independent domains:

- The World, that is the agent's external environment, comprising both the physical world (roads, obstacles, weather) and the social world (other agents, traffic rules, institutional constraints).
- The Self, that is the agent's internal environment, consisting of its concrete implementation: the hardware, software, and knowledge it uses to guide its decisions.

The agent receives inputs from the World, unstructured data such as sensory information or text, and produces outputs that can impact both the World and the Self. Inputs can be physical, such as sensor readings, or social, such as messages from other agents.

An agent is a dynamic system whose behavior can be defined by a set of states that change as actions are performed. Each state has two components: one that reflects the agent's perception of the World, and one that represents the state of the agent's Self.

Actions fall into two categories. Some actions are controllable: the agent chooses to perform them. Others are uncontrollable: they are performed by the environment and impact the agent's state regardless of its choices. This distinction is fundamental to understanding the interaction between an agent and its environment. In game-theoretic terms, the agent and the environment are players in a game, each with their own actions and goals. The agent controls its own actions but must anticipate the environment's responses. This is why planning by anticipating future states is essential for autonomous agents.

In chess, the agent controls its own moves but cannot control its opponent's moves. A self-driving car controls its speed and steering, but it cannot control the behavior of other vehicles, pedestrians, or weather conditions. When an action is executed, whether chosen by the agent or imposed by the environment, the agent moves from its current state to a new state. Both the agent's internal state and its perception of the world may change as a result.

A goal is a property of the agent's behavior that must be achieved during its interaction with the environment. To achieve a goal, the agent must develop a plan by anticipating how the environment will react to its actions and choose a course of action that leads to success.

Goals fall into two categories.

**Invariant goals** are constraints that must be maintained continuously. They are never fully "achieved" in the sense of being completed; rather, they must always be respected. The agent continuously monitors and enforces them. Safety is the classic example: an autonomous car must never collide with anything, regardless of what else it is doing. Other examples include keeping resource usage within reasonable limits, ensuring performance exceeds a certain threshold, or always maintaining a positive bank account balance.

**Transient goals**, by contrast, guide the agent toward a specific target. Once that target is reached, the goal is achieved and can be set aside. Examples include completing a parking maneuver, responding to a message, or charging the battery to 80%. Transient goals are what drive the agent's day-to-day activity. They are generated in response to the current situation and are pursued until they are achieved or abandoned.

These two types of goals are complementary. Invariant goals prevent undesirable outcomes, but they do not ensure progress. Transient goals ensure progress, but they do not guarantee that undesirable outcomes are avoided. This is why they must be pursued jointly. An autonomous car drives toward its destination (a transient goal) while continuously ensuring safety (an invariant goal).

The set of goals that an agent can pursue characterizes the scope of its mission and its degree of autonomy. A car equipped with simple lane-keeping has limited autonomy. A car that can handle overtaking, intersections, and complex traffic scenarios has far greater autonomy. The richer the set of goals, the greater the agent's autonomy.

### *2.2.2 The Agent's Faculties: The Tools of Agency*

The agent faculties are the means by which an agent carries out its mission: the resources, skills, and powers at the agent's disposal. Without the necessary faculties, an agent may certainly know what needs to be done, but may not have the material and intangible resources required to accomplish it. A police officer has the authority to make arrests. A self-driving car is certified to operate on highways. A chatbot has access to a knowledge base.

We distinguish four types of agent faculties.

**Capabilities** determine whether a goal or action is physically or procedurally feasible in a given context. These include physical capabilities, such as acceleration and deceleration rates for autonomous vehicles; professional capabilities, such as being a doctor; and certified capabilities, such as holding a license. Certified capabilities are particularly important for autonomous systems, as they provide a basis for trust and accountability. When an agent holds a certification, such as a driver's license for a self-driving car, it signals that the agent has been evaluated against established standards.

**Competencies** are the reasoning powers that make an agent effective. They enable it to think through problems, weigh alternatives, and make decisions. A chess-playing agent has competencies for evaluating board positions. A self-driving car has competencies for predicting the behavior of other vehicles. Without competencies, an agent might have the physical ability to act, but it would not know how to act wisely.

**Aptitudes** allow effective communication and synergy. They include making requests, negotiating, and building consensus. Agents with strong aptitudes can make requests of other agents, understand complex situations, take the initiative in negotiations, and reach agreements. Aptitudes are the social skills of agency. They enable agents to interact effectively with other agents, to make requests, negotiate, build consensus, and resolve conflicts.

**Authority** defines the power relationships between the agent and other agents in the system. This includes the agent's roles, such as orchestrator, judge, director, administrator, and its membership in a group of agents enjoying specific privileges. Authority is the institutional dimension of agency. It defines what the agent is permitted to do, what it can require of others, and how it fits into the hierarchy of the autonomous system.

There is no shortage of examples illustrating how agents are differentiated based on their faculties. Any autonomous car should have the minimum capabilities required to drive safely and should eventually be certified in accordance with applicable standards. A police car, moreover, is authorized to stop vehicles and enforce traffic laws. A broker should possess the necessary competencies to effectively analyze situations and make decisions. Agents should also have the aptitudes to negotiate and reach a consensus in order to close deals. Finally, they should have the authority to execute transactions within predefined limits.

#### *2.2.3 Fitting In: Integration Conditions*

No agent operates in a vacuum. A car cannot drive through a wall. A police officer should not ignore traffic laws. A chatbot should not insult its users. These are the constraints that shape behavior: the physical, social, and normative conditions that every agent must respect to function effectively in its environment. Integration conditions are the rules of the game: they define what is possible, what is permitted, and what is desirable.

The physical environment imposes hard, inviolable limits on what the agent can do. A road blocked by concrete barriers prevents a vehicle from moving forward. These are physical constraints: they are inviolable. The social environment, by contrast, restricts the agent's freedom through norms, rules that are enforceable but not absolute. Traffic lights indicate what is permitted or prohibited without ever physically blocking the way. Norms can be violated, but such violations may result in sanctions. The law not only sets forth prohibitions but also specifies the penalties for violating them.

The agent's integration conditions restrict its behavior for harmonious integration into its environment. They are of three types.

**Physical constraints** are conditions that the agent's state must always satisfy. If a physical constraint is violated, the agent is in an inconsistent state, which is simply not possible in its environment. These constraints are inviolable. They ensure that the agent's behavior is grounded in the physical world, that its actions are consistent with the laws of physics. This is why the grounding problem discussed later is so important: an agent whose internal model violates physical constraints cannot be trusted to act safely in the real world.

**Norms** are rules that determine, for a given situation, which of the possible actions are permitted, required, or prohibited. They include rules derived from the application of the law or ethical principles. Failure to comply with these rules may result in sanctions according to an associated value system discussed below. Typical examples of norms are traffic rules, priority rules for sharing resources and managing tasks, as well as ethical rules.

**Value systems** allow the agent to evaluate the expected utility of an action by balancing costs, benefits, and risks. Value systems operate at the level of practical knowledge. They guide action by assigning utilities to different outcomes, expressing preferences and priorities. While norms define what is permitted or prohibited, value systems define what is preferred; the two work together in decision-making. The cost of using new resources may result in a performance gain. A violation of norms may lead to performance gains but also to sanctions, a classic trade-off. Value systems help the agent choose between competing goals or resolve conflicts between actions.

Reaching the same destination can be achieved by train or by plane, with different costs in terms of time and money; the value system helps the agent decide. For an autonomous vehicle, the value system might weigh safety, comfort, efficiency, and legality. When these values conflict, for example, when the safest route is also the slowest, the value system provides the basis for making a trade-off. Safety might be given the highest priority, with comfort and efficiency weighted according to passenger preferences.

#### *2.2.4 Staying Alive: The Agent's Operating Conditions*

An autonomous agent must ensure that a set of operating conditions, which are essential to the fulfilment of its mission and depend on its internal states, are met. It requires fuel, maintenance, and, at times, the cooperation of other agents. If any of these conditions are not met, the agent proactively takes steps to remedy the situation by pursuing the appropriate goals. The agent's proactive behavior is essential to guarantee its autonomy by taking care of itself, rather than merely reacting to the outside world.

The distinction between a goal and an operating condition lies in the fact that the latter describes the internal states that must be maintained for the agent to accomplish its mission. The car needs a battery charge level above a certain threshold, brakes in good working order, and operation free of software glitches. These are not goals, but prerequisites for the car to pursue its goals.

We have already discussed what the agent is supposed to accomplish (its mission and goals), the resources at its disposal (its faculties), and the constraints it must adhere to (its integration conditions). We will now focus on what enables it to remain alive and operational: its operating conditions.

There are four types of operating conditions.

**Survival conditions** concern the availability of resources needed for the system's vital functions. These include energy, memory, computation, communication, physical integrity, and environmental tolerances such as temperature and pressure. When a survival condition fails, the agent must act proactively to restore it.

**Safety and security** conditions address technical risks related to failures or malicious actions that affect the agent's internal operation. They enable the detection of hardware or software failures, as well as threats related to malicious actions. Malicious actions relate to data integrity, access control, and built-in security mechanisms such as backup and redundancy.

When a safety or security condition is not met, for example, in the event of a software failure or a detected intrusion, the agent must take actions to restore internal integrity.

**Self-fulfilment conditions** refer to the agent's ability to adapt and improve. This includes learning and acquiring skills, reconfiguring resources, and maintaining the ability to operate in a changing environment. Self-fulfilment is about moving from common empirical knowledge (what the agent currently knows) to higher levels of knowledge, integrating new experiences, updating models, and improving competencies. This is the most challenging type of operating condition, as it requires the agent to recognize its own limitations and take action to overcome them.

When these conditions are not met, for example, when the agent's competency model becomes obsolete, the agent must take proactive steps to update its capabilities. Consider the example of an autonomous vehicle whose navigation system relies on a map that is several years old. While the vehicle can certainly continue to operate, its ability to navigate effectively in a constantly changing urban environment is compromised. The agent must identify this shortcoming and define a meta-goal aimed at updating its map data. Similarly, an agent may need to reconfigure its resources and how they are managed to adapt to changes in its environment.

**Social recognition** conditions indicate that the agent is accepted by others in accordance with its faculties related to interactions with other agents. When social recognition fails, for example, when other agents do not trust the agent's information or do not respond to its requests, the agent must act proactively to restore its social standing, such as providing evidence of its reliability or adjusting its communication style.

To illustrate how social recognition conditions work in practice, consider a delivery robot operating in a smart factory. The robot has the operating condition "other agents accept my delivery assignments as reliable." This is not a goal; it is a precondition for the robot to function effectively in the social environment of the factory.
If the robot's sensors malfunction, causing it to deliver several packages to the wrong locations, other agents—the warehouse management system, the receiving robots, and the human supervisors—may lose trust in the robot's reliability. The robot is no longer accepted as a dependable member of the system. Its social recognition condition is violated. The robot detects this violation and generates a set of goals that can restore its social recognition. The candidate goals might include: request a diagnostic check of its sensors, offer to perform a series of test deliveries to demonstrate reliability, or request human verification of its future deliveries. The robot evaluates each option against its value system, considering time, energy cost, and the urgency of pending deliveries, and selects the most advantageous one.
This example illustrates how all the features we have discussed, such as goals, integration conditions, faculties, and operating conditions, work together in practice. The agent monitors its state, detects when an operating condition is violated, and acts proactively to restore it. This is the essence of autonomous behavior: the agent does not merely pursue goals; it also ensures that the conditions for goal pursuit remain intact.

**Summary: The Agent's Knowledge Attributes**

We summarize below the four knowledge attributes and their subcomponents:

| Attribute | Subcomponents | Description |
|---|---|---|
| **Mission** | Invariant goals, Transient goals | What the agent is supposed to achieve |
| **Faculties** | Capabilities, Competencies, Aptitudes, Authority | The means at the agent's disposal |
| **Integration Conditions** | Physical constraints, Norms, Value systems | Constraints for harmonious integration |
| **Operating Conditions** | Survival, Safety/security, Self-fulfilment, Social recognition | Conditions for remaining operational |

***2.2.5 Agent Knowledge Attribute Specialization: Two Examples in Brief***

Having defined the four knowledge attributes, we now illustrate how they are instantiated for two agents depending on their mission.

**The Autonomous Driving Agent**

The autonomous driving agent operates in the physical world. Its mission is to transport passengers safely and efficiently from one location to another. This mission requires a knowledge base that includes maps, traffic laws, vehicle dynamics, and behavioral models of other road users.

The autonomous driving agent's capabilities are primarily physical: the vehicle's speed, braking capacity, range, and set of sensors. These capabilities define what the car can physically do that is, its scope of action. Its capabilities include the ability to predict the behavior of other vehicles and to plan safe trajectories. Its scope of action is limited to what the vehicle is certified to do (for example, driving on the highway or in the city).

The integration conditions of the autonomous agent are determined by the laws of physics, traffic laws, and the expectations of passengers and other road users. A car cannot violate the laws of physics: it cannot accelerate faster than its mass and engine power allow. It must obey traffic laws, such as stopping at red lights, yielding to pedestrians. And it must meet passengers' expectations regarding comfort and safety.

Finally, the autonomous agent's operating conditions are tied to its internal state: battery level, tire pressure, and the condition of its sensors and software. If the battery drops below a critical level, the agent's proactive module must generate a meta-goal to recharge. If a sensor fails, it must adapt or be repaired. These operating conditions are essential to the agent's ability to fulfil its mission over the long term.

**The Communicator Agent**

The Communicator operates in the social domain of information exchange. Its mission is to exchange information reliably and accurately with other agents. This mission requires a knowledge base comprising communication protocols, semantic linguistic models, norms, and trust models for other agents.

The Communicator's faculties are primarily linguistic, in particular the ability to interpret messages in context and adherence to protocols. Its competences include the ability to detect deception, negotiate, and build consensus. Authority is defined by the agent's role in the communication hierarchy (e.g., moderator, spokesperson, participant).

The conditions for the Communicator's integration are determined by communication norms. Communication protocols determine communication norms and are akin to physical constraints, such as bandwidth limits, latency constraints, and message formats. Norms, such as politeness, sincerity, and relevance, are akin to communication rules.

Finally, the communicator's operating conditions pertain to its internal state: buffer space, network connectivity, and processing capacity. If buffer space runs out, the agent must act proactively to delete old messages or request a reduction in data throughput. In the event of a loss of connectivity, it must adapt or attempt to re-establish the connection. These operating conditions are essential to the agent's ability to fulfil its mission over the long term.

**Summary**

These two examples show how the same faculty attributes take different types of values depending on the domain. The driving agent's capabilities are physical; those of the communicator are linguistic. The integration conditions for the driving agent are governed by physics and traffic laws; those for the communicator are governed by protocols and standards. Yet we will show that both agents can share the same underlying generic architecture.

## 2.3 Agent Reference Architecture: Specification and Behavior

The use of reference architectures is very common in systems engineering, particularly for complex systems. An architecture provides a constructive solution to a problem: a blueprint that can be tested, refined, and reused. Much like an architectural plan for a building, it shows the structure and relationships between components, but does not specify every detail of the construction. It is a guide for implementation, not the implementation itself.

In the previous section, we identified the agent's knowledge attributes. We will now explain how the agent's attributes, which are part of its knowledge, are used by the architecture's modules to generate behavior consistent with its mission.

The proposed architecture is a theoretical model designed to integrate all the features of artificial agents in order to more closely approximate the behavior of a human agent. It generalizes the architectures used in robotics and seeks to cover all aspects of agency. It provides a basis for reflection on agent design, as well as a common framework for understanding and comparing different agentic architectural solutions. It does not prescribe a specific implementation; rather, it identifies the functional components that every autonomous agent must possess, as well as the relationships between them (Figure 1).

As explained earlier, an agent is a dynamic system that receives inputs from its environment, composed of the World and the Self, and produces outputs intended to modify the state of that environment in order to accomplish its mission.

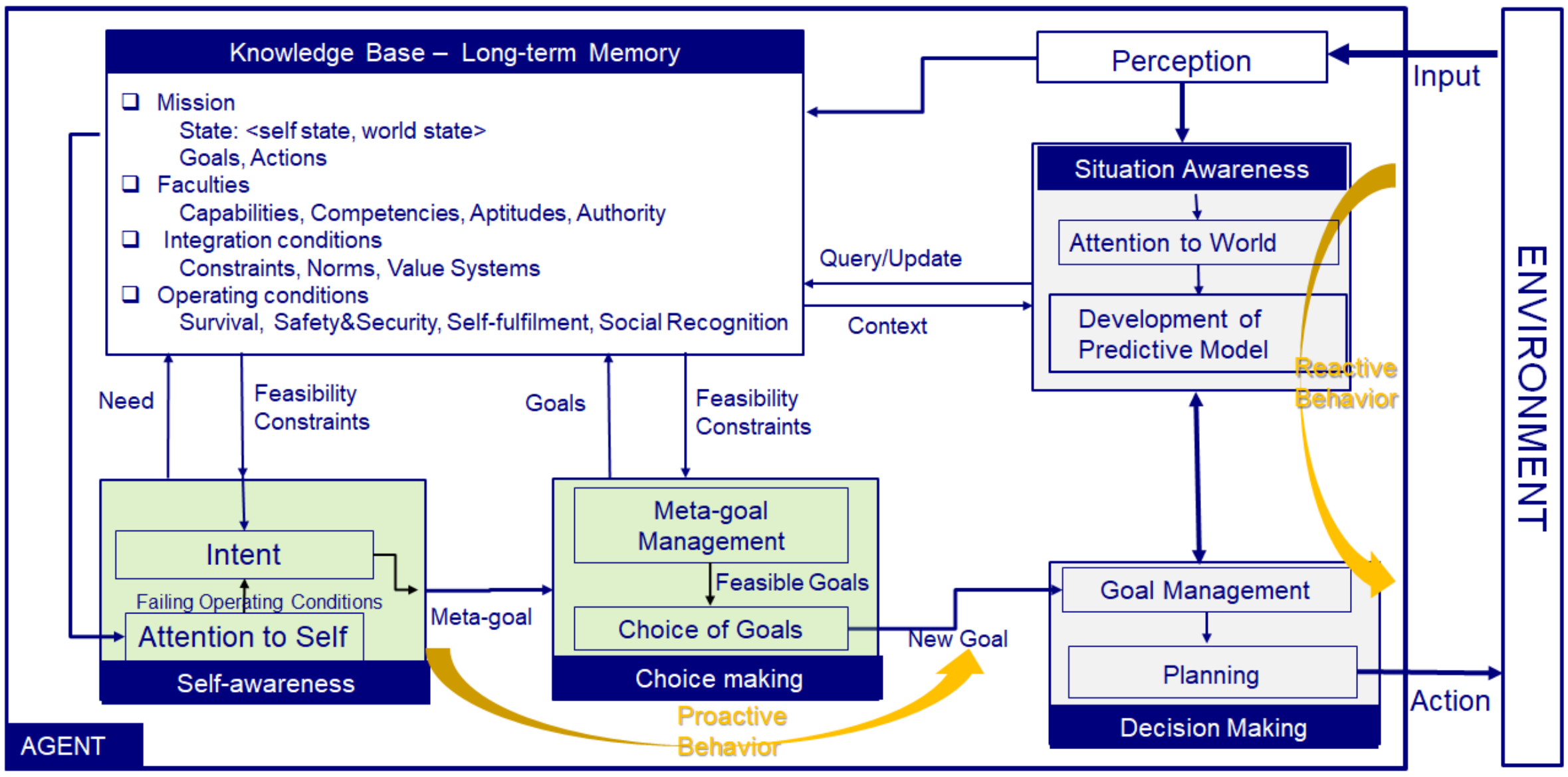

Figure 1: The agent's architecture, organized around its knowledge base.

The knowledge base serves as the agent's memory. It contains everything the agent knows about its mission, including its state, goals, and actions. It also contains specifications of its knowledge attributes, as defined in the previous section.

As explained earlier, the agent's state consists of two components: the state of the World, which represents the perceived state of the agent's external environment, and the state of the Self, which represents the internal state of the agent itself.

The architecture consists of three modules, each performing a distinct function. Their coordinated execution enables the agent to pursue its goals and accomplish its mission. The knowledge base provides the three modules with the knowledge that guides their decisions. In turn, the modules update the knowledge base to keep it consistent with their decisions.

**The Perception Module**

The perception module collects raw data, such as images, sounds, text, sensor readings, from the World and transforms it into a coherent representation of the state of the World.

The perception model provides a solution to the framing problem discussed in the following section: How does the agent know what is important in a given situation? It uses the knowledge base to determine the context, which allows it to focus on relevant information and ignore irrelevant details. It verifies that the perceived information is consistent with the agent's integration constraints, then either corrects it or triggers an alert. After this verification, it updates the World state component.

**Example:** The perception module of an autonomous car detects a red traffic light ahead. It identifies the color of the light, estimates the distance to it, and verifies that the car is approaching an intersection. This perceived information is then stored in the World state to be used by the reactive module.

**The Reactive Module**

The reactive module describes an agent's response and adaptation to changes in the World. It implements a cyclical reactive process triggered by these changes, which are detected by the perception module, and involves two functions.

**Situation Awareness:** This function develops a predictive model of the World based on the agent's mission goals. First, based on the perceived World state, it must analyze what this implies for the agent's goals; for example, a traffic light turning red or a pedestrian stepping onto a crosswalk. It examines the actions it could take by consulting the agent's knowledge base, including its driving capabilities and the applicable traffic rules. It then develops the predictive model, which takes into account possible reactions from the environment: a map of possible futures arising from the current state, much like a chess player anticipating several moves ahead.

**Decision Making:** This function defines a course of action. Based on the predictive model, it identifies the goals applicable to the agent's current state. It then generates a plan to achieve these goals; that is, a set of sequences alternating between controllable and uncontrollable actions that enable the agent to achieve its goals. The planning process is constrained by the agent's knowledge: its capabilities (what it can do), the applicable norms (what it is permitted to do), and the associated value systems (what it prefers to do). Based on this knowledge, the function selects a sequence and executes the first controllable action. After execution, the function updates the state of the World. This completes the cycle of the reactive process.

In the following section, we discuss in greater detail the challenges involved in implementing the reactive module.

**The Proactive Module**

While the reactive module responds to external changes in the World, the proactive module monitors the agent's internal state, the Self, and continuously verifies the validity of its operating conditions stored in the knowledge base. It reflects the agent's ability to adapt to changes in its internal state and to meet the needs essential to the fulfilment of its mission. It involves two functions.

**Self-Awareness:** When an operating condition becomes invalid, this function determines a meta-goal; that is, a set of possible goals that could restore that condition. An operating condition can be a simple state predicate, a complex condition dependent on the values of key performance indicators (KPIs), or even conditions generated by AI monitors to predict potential system failures.

The meta-goal associated with an operating condition can be either pre-calculated and stored in the knowledge base, or defined by sets of rules, or even determined through machine learning. The intent to correct a malfunction arises only if the meta-goal is not empty. For example, if the battery is low, the meta-goal might include: recharge at station A, recharge at station B, or replace the battery.

**Choice Making:** For a given meta-goal, this function performs an analysis to select the best option, always taking into account the knowledge contained in the knowledge base. In particular, it checks which of the goals of the meta-goal are feasible, which are permitted by the norms, and which offer the greatest utility according to the applicable value systems. The selected goal is transmitted to the reactive module for execution.

**Reactive Behavior vs. Proactive Behavior**

Most agent architectures support only reactive behavior. This is sufficient for simple tasks, but not for full autonomy. Proactive behavior, that is, the ability to anticipate operational problems and take the initiative, is essential for agents that must operate reliably over the long term.

The two modules are complementary: reactive behavior responds to the World; proactive behavior ensures that the agent can continue to respond to the World by maintaining the health of the Self. Without proactive behavior, an agent risks exhausting its resources, failing to adapt, or losing the trust of other agents. Without reactive behavior, an agent cannot respond to opportunities and threats in its environment. Both are essential for long-term autonomy.

This distinction is familiar to us from our human experience. Reactive behavior corresponds to what we do during a conversation or while walking: automatic and immediate responses to external stimuli. Proactive behavior is driven by needs originating in the Self. When I'm hungry, I can order a meal, cook at home, or go to a restaurant. This choice depends in part on my personal circumstances (physical condition, transportation options, bank account) and on external conditions (weather, accessibility, and availability of services). Furthermore, my choices may be influenced by norms and value systems that allow me to weigh the cost of a choice against the expected satisfaction.

## 2.4 Agent Reference Architecture Specialization: Two Examples

### *2.4.1 How the Self-Driving Agent Works*

The Self-Driving Agent's behavior follows the same three-process pattern as the reference architecture, with concrete physical instantiations.

**Perception: Interpreting the World**

The perception module takes in a flood of data, such as camera images, radar signals, lidar scans, and transforms it into a coherent picture of the World. The module collects raw sensory data, analyzes it, and verifies its consistency against constraints from the knowledge base. Does the new information make sense? Is the pedestrian in a plausible location? Is the traffic light consistent with the car's position? If something does not add up, the car raises a warning or discards the data. Otherwise, it updates the World state.

For example, if it perceives a red traffic light ahead, it identifies the light's color, estimates its distance, and updates its World state. It checks that it is approaching the intersection and that the distance to the stop line is feasible.

**Reactive Behavior: Responding to the World**

When the World state is updated, the car must decide what to do, starting a cycle of the reactive process.

**Situation Awareness**: First, the function analyzes the state of the surrounding environment to identify obstacles around the car determining their positions and kinematic characteristics, as well as traffic signs and key elements of the roadway. It detects a red light ahead. It spots a child on the crosswalk. It measures the distance between itself and the car in front of it.

Next, it develops a predictive model by estimating how the external environment will react to possible controllable actions, based on the agent’s knowledge attributes.

**Decision Making:** The agent analyzes the predictive model to determine the set of applicable goals, such as braking, turning, accelerating, or signaling, while respecting the rules of the road and the laws of physics. Running a red light is off the table. Driving onto the sidewalk is not an option.

Next, for the relevant goals, the planner consults the integration conditions stored in the knowledge base to generate a corresponding plan that defines the sequences of permitted actions. If the first controllable actions conflict—should the driver brake abruptly or turn the steering wheel gently?—the agent resolves this conflict using its value system, which is also stored in the knowledge base. Safety always takes precedence over comfort. Executing the action updates the state of the World and ends the reactive behavior cycle.

For example, if the car approaches a stop sign, applicable goals include "stop before the stop line." Planning yields a braking action. The scheduling function selects a smooth, comfortable deceleration. Execution brings the car to a complete stop before the line.

**Proactive Behavior: Staying Safe and Functional**

**Self-Awareness:** If an operating condition is not met, for example, if the battery level drops below 15%, the agent generates a meta-goal: to restore that condition. The car determines the possible options based on the agent's knowledge attributes: recharge at Station A, recharge at Station B, or return to base.

**Choice Making:** The agent selects the most advantageous goal for the current state after consulting the capabilities and integration conditions stored in the knowledge base: which ones are permitted by the standards? Which ones match the car's capabilities? Which ones minimize delay or maximize energy efficiency? The selected goal is transmitted to the reactive module to be executed as a transient goal.

### *2.4.2 How the Communicator Agent Works*

The Communicator sends messages, receives them, and, most importantly, interprets what they mean. It knows when to be polite, when to be direct, when to ask a clarifying question, and when to give a definitive answer. It operates in the social World, where words matter as much as actions. Its behavior is organized around the same three processes: perception, reactive behavior, and proactive behavior.

**Perception: Interpreting the World**

When a message arrives, the Communicator must first understand what it means. It makes a semantic analysis of the message to identify who sent it, who it is for, what type of message it is, and what it says. More importantly, it asks: what does the sender actually want? Is this a question, a command, or a statement of fact?

Then, it checks for problems. It applies the constraints from the knowledge base to verify consistency. If an anomaly is detected, the message is rejected or flagged for review. Messages requiring a response are forwarded to the reactive module.

Finally, it updates its memory. It adds the message to the queue, logs the conversation, and stores any relevant information in the knowledge base, including the World state (the state of the communication environment) and the Self state (its internal resources).

**Reactive Behavior: Responding to Messages**

When new messages arrive, the Communicator must decide how to respond.

**Situation Awareness:** Based on the Communicator's state, which is updated, in particular, by the Perception module, the Reactive module identifies the applicable goals. If someone asks a question, the goal is to answer it. If someone gives an order, the goal is to comply with it, provided it is legitimate. It consults the knowledge base for applicable goals and norms. It also assesses the context: Who is the sender? What is their relationship to the agent? What are the expectations in this situation?

Then, it constructs a predictive model. It considers possible responses and anticipates how the sender might react to each. Will a direct answer be appreciated, or would a more diplomatic response be appropriate? Does the situation require immediate action or can it be deferred?

**Decision Making:** Based on the predictive model, the agent plans its response. It considers possible actions, such as sending a reply, forwarding the message, or asking for clarification, while respecting the norms of communication stored in the knowledge base. If multiple messages compete for attention, the Communicator uses its value system to decide which to handle first. Urgent messages take priority. High-bandwidth tasks are deferred if the network is congested.

When it executes the selected communication action, it sends the reply, logs the interaction, and updates its internal state in the knowledge base.

**Proactive Behavior: Staying Healthy**

**Self-Awareness:** The proactive module monitors the Communicator's operating conditions. It asks itself: "Is everything okay? Can I continue?". If an operating condition is false, for example, buffer free space is below 10%, the agent generates a meta-goal. This is a set of candidate goals that could restore the condition: clear the cache, increase buffer size, request retransmission, or switch channel.

**Choice Making:** For a given meta-goal, the agent chooses the best one by consulting the knowledge base. Which goals are permitted by norms? Which resources are available? Which align with its communication faculties? The selected goal is transmitted to the reactive module for execution as a transient goal.

For example, if the buffer free space is below 10%, the candidate goals of the meta-goal can be: clear old messages, increase buffer size, or request throttling from other agents. If the feasibility assessment discards "increase buffer size" because no memory is available, the goal is discarded. If "clear old messages" is the best option, the goal is added to the pending goals of the reactive module, which deletes the oldest messages, freeing buffer space.

## 2.5 Our Contribution

The architecture we present identifies the functional components that every autonomous agent must possess, as well as the relationships that connect them through an agent knowledge base.

Our architecture shares conceptual ground with the BDI (Belief-Desire-Intention) model of agency [9][10], which has been widely adopted as a framework for practical reasoning agents. However, it differs from the BDI model in several fundamental respects.
First, while BDI agents rely on a pre-defined plan library from which plans are retrieved and executed [11], our architecture computes plans online in interaction with the environment, enabling adaptation to unforeseen situations.
Second, perception is not explicitly modeled in the BDI framework, since it is assumed that beliefs are updated by external events that are simply presented to the agent.
Third, BDI agents are primarily reactive to events, whereas our architecture includes a proactive module that continuously monitors the agent's internal state and generates meta-goals to maintain operating conditions essential for long-term autonomy.
Fourth, normative reasoning is central to our architecture as integration conditions constrain planning and decision-making, whereas it is not a core component of the standard BDI model [12].

Our architecture participates in a broader movement that integrates AI modules with structured knowledge and memory, while addressing fundamental limitations in existing models.

**Neurosymbolic AI** approaches, as systematized by [13], aim to combine the perceptual strengths of neural networks with the reasoning capabilities of symbolic AI. This integration draws inspiration from the human cognitive system, where perception (System 1) is complemented by deliberate reasoning (System 2) based on explicit knowledge [13]. While neurosymbolic AI provides a powerful paradigm for linking data-driven perception with symbolic knowledge, it often remains focused on the reasoning and learning components. In contrast, our architecture provides a comprehensive framework for agency that explicitly integrates these components with perception, decision-making, planning, and proactive self-maintenance, within a complete agent architecture.

Concurrently, approaches leveraging **world models** for planning [14] propose to equip large language models with an internal world model to simulate state transitions and perform deliberate planning. The framework of [14] uses a world model to predict the outcomes of actions and guide reasoning. This aligns with our architecture's emphasis on building predictive models during the reactive module's situation awareness phase. However, the world-model approach typically treats planning as a computational process within a learned model, whereas our architecture also incorporates normative reasoning and proactive self-maintenance as explicit architectural components that shape goal selection and planning.

From the perspective of **autonomic and self-aware computing**, our architecture extends principles established by IBM's Autonomic Computing initiative [15] and the broader field of self-aware systems [16]. The autonomic computing vision of self-configuring, self-healing, self-optimizing, and self-protecting systems aligns with our proactive module's function of monitoring operating conditions and generating meta-goals to maintain system integrity. However, these earlier frameworks largely focused on system-level properties and resource management, whereas our architecture provides a cognitive foundation for autonomous agency in physical and social environments.

**Agent-oriented software engineering** [17] provides methodologies and languages for developing multi-agent systems, but often assumes that agents operate with pre-defined capabilities and interaction protocols. Context-aware multi-agent systems [18] extend this by incorporating environmental information into agent decision-making, yet they typically lack the integrated cognitive architecture that we propose.

Our architecture synthesizes and extends these strands. Like neurosymbolic AI, it integrates perception with knowledge-guided reasoning. Like world-model approaches, it performs online planning through interaction with the environment. Like autonomic systems, it includes proactive self-maintenance. Our architecture extends these specialized models and provides a unified framework that integrates their capabilities into a single, coherent architecture organized around a knowledge base. This synthesis makes it a comprehensive reference model for autonomous agents that must operate reliably in dynamic, uncertain environments subject to physical and social constraints.

## 3. From Architecture to Implementation: Key Challenges

The architecture presented here provides a principled conceptual framework for understanding autonomous agents and characterizing their behavior. However, there is a gap between the principles of structuring and coordination and their actual implementation. This section addresses the key questions that arise when creating concrete agents: how to link unstructured perception to structured knowledge; how to make safe predictions about interactions with a dynamic environment; how to plan actions to achieve goals; and how to manage the complexity of systems engineering that arises when these solutions are integrated. We address these problems in turn.

### 3.1 Linking Unstructured to Structured Data – RAG Architecture

Perception is the agent's first challenge. The World generates a stream of raw sensory stimuli that the agent must interpret. This is hard enough for a single agent. For a multi-agent system, it is even harder: each agent must not only understand its own World but also communicate that understanding to others.

The complexity of perception stems from several factors. One is the difficulty of correctly analyzing data and extracting relevant information from it. This difficulty is linked to the ambiguity of the input data (which can give rise to different interpretations) or to its imprecision (blurred or noisy stimuli). Furthermore, this type of complexity is compounded by the volume of raw data required to extract relevant information under real-time constraints.

Another notable challenge is linking the extracted information to the agent's internal representations, stored in its knowledge base. To do this, the agent must rely on a semantic similarity relationship that is insensitive to noise or ambiguity. A landscape remains fundamentally the same, whether it is covered in snow or subject to variations in lighting. We can understand the meaning of a sentence expressing the same idea, even if it uses different words or grammatical structures.

This interpretive ability distinguishes agents from traditional automated systems that operate on structured, predefined data domains. Its implementation raises what is known as the framing problem: how does the agent know what matters in a given situation? The framing problem, originally identified in AI by McCarthy and Hayes [19], captures the difficulty of representing the relevant aspects of a situation without explicitly enumerating all possible irrelevant facts. It is the problem of determining which information is salient and which can be safely ignored; this problem becomes acute when the agent operates in an open, dynamic environment.

This problem manifests itself in a very concrete way in RAG (Retrieval-Augmented Generation) architectures, which are one of the cornerstones of most multi-agent architectures. The RAG framework, introduced by Lewis et al. [20], integrates an LLM with a long-term memory containing domain-specific knowledge, in order to semantically enrich and control the LLM's responses.

The principle behind RAG architecture is illustrated in Figure 2, where an LLM receives queries phrased in natural language, the answers to which may extend beyond the scope of its general knowledge. For example, queries may concern a company's financial situation or a store's inventory levels on a given date. The idea is that this domain-specific data is stored in a memory from which the LLM must extract all relevant information before formulating its response.

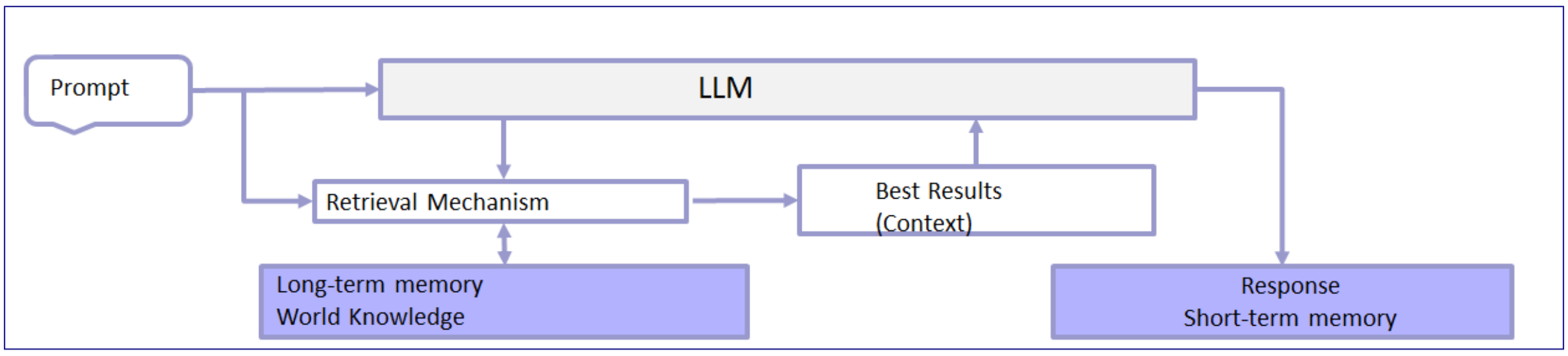

Figure 2: A RAG architecture.

The practical implementation of RAG architectures poses a series of technical challenges, all of which are manifestations of the framing problem:

- How can a large volume of knowledge be represented and managed in memory? This is the problem of knowledge representation: how to structure knowledge so that it can be retrieved and used effectively. This knowledge must be structured in a way that allows for reuse based on semantic principles, as it must be updated regularly, and inconsistencies must be avoided. At present, we do not have satisfactory structuring and representation methods that meet the requirements of semantic fidelity and efficiency.

- How can we extract relevant knowledge from memory based on various similarity relationships, including perceptual, semantic, thematic, and functional ones? This is the problem of relevance: given a database and its data, how can we extract all relevant information related to a query?

Two broad approaches have emerged to address these challenges.
The first is **embedding-based retrieval**, where knowledge is represented as dense vector embeddings and retrieval is performed using similarity metrics such as cosine similarity [21]. This approach is efficient and scalable, and it handles unstructured data well. However, it sacrifices precision and interpretability, since the agent cannot justify why a particular piece of knowledge was retrieved or verify its correctness [22].
The second is **knowledge graph-based retrieval**, where knowledge is represented as a graph of relationships between entities, and retrieval involves traversing the graph using structured queries. This approach provides enhanced precision, interpretability, and support for logical reasoning and inference [23]. However, it sacrifices efficiency and scalability as the graph grows, and effective retrieval requires the ability to formulate precise queries, which presupposes that the agent already knows what matters, exactly the framing problem.

Neither of these two approaches is satisfactory for autonomous agents. Embedding-based search offers efficiency and scalability, but at the expense of accuracy and interpretability. Conversely, knowledge graph-based search offers accuracy and interpretability, but at the expense of efficiency and scalability. Hybrid approaches attempt to combine the two [24][25], while introducing orchestration complexity and without fundamentally resolving the tension between semantic fidelity and search efficiency. What is needed is a unified framework that integrates knowledge representation and retrieval, allowing the agent to dynamically determine what matters based on its mission, goals, norms, and current state, a solution to the framing problem that remains unresolved [26].

Take, for example, a query from a video ringtone service subscriber addressed to the provider: *"Hello, my name is Paul Brown, customer ID xx. My friends told me they couldn't see my video when they called me. Can you check if my video ringtone service is active? If it's inactive, please activate it for me and set 'City Night Aerial' as my ringtone."*

To fulfil this request, the system must break it down into simple operations on the data in memory. This requires several capabilities that current systems lack: semantic parsing (understanding the user's intent from natural language), task decomposition (breaking the request into steps), knowledge retrieval (finding the relevant data), and action execution (generating a response and updating the database). Each of these steps carries a risk of failure. It must verify whether the customer and all relevant data are in memory, check the status of their video ringtone service, and, if it is inactive, activate it. Then, if "City Night Aerial" is among the available ringtones, the system will update the customer's preferences.

Current solutions to this problem remain largely unsatisfactory. The current accuracy of “text-to-SQL” translations by language models is generally less than 90 percent, although this figure varies depending on the performance tests [27][28]. This is largely unsatisfactory. With a success rate of 0.9 per translated query, for 10 queries, the overall reliability is $0.9^{10} = 0.3487$, less than 35%! This calculation reveals the gap between current capabilities and the requirements of autonomous systems that must operate without human intervention for extended periods, where even a 10% per-operation failure rate compounds rapidly.

It should be noted that our architecture is based on the RAG paradigm, as the various cognitive functions must make extensive use of the knowledge stored in the agent's knowledge base.

### 3.2 Awareness of the Physical and Social World

In our reference architecture, we have assumed that an agent is aware of the constraints imposed by its environment and of its own capabilities. To effectively utilize this knowledge, the agent must evaluate these constraints and draw conclusions regarding the consistency of its state, the controllable actions that are permitted, required, or prohibited, and anticipate the environment's possible reactions to these actions.

Situational awareness involves reasoning and decision-making capabilities. In terms of the agent architecture, it is the responsibility of the reactive module, which constructs a predictive model of the environment based on the agent's mission goals. The quality of this model depends on the quality of the knowledge in the knowledge base and the agent's ability to update it. It assumes that the agent's state accurately reflects what is happening in the environment. Indeed, partial knowledge of an ambiguous environment must be supplemented with information from the knowledge base as accurately as possible.

The complexity of this reasoning illustrates the difficulty of constructing a precise and predictive semantic model of the environment. A traditional automated system operating on structured data does not face this problem, because its inputs are unambiguous. The ambiguity is a consequence of operating in the non-technical world.

For example, in some autonomous vehicles, reconstructing the state of the environment may combine perceptual information with pre-recorded, semantically rich maps detailing road infrastructure to build an even more accurate semantic model of the environment.

The complexity of knowledge management has two aspects.

The first involves promoting situational awareness by providing the appropriate knowledge. For example, if an obstacle has been identified, the reactive module must receive information from the knowledge base regarding its relevant properties in order to enrich its semantic model; or estimates must be calculated for the parameters used in the decision-making process, such as travel times, traffic density, and risk parameters.

The other aspect involves discovering new knowledge necessary to deal with entirely new situations not anticipated during design. This corresponds to the self-fulfilment operating conditions: the agent must adapt and improve its capabilities in response to changing circumstances. This is the most challenging type of adaptation, as it requires the agent to recognize gaps in its knowledge and take action to fill them. Situational awareness can be improved through the discovery of new concepts, or decision-making can be enriched by new goals that adapt to a changing environment. This is the problem of learning and adaptation, because the agent must not only use its existing knowledge but also expand it as it encounters new situations.

To address these issues, the agent must be able to manage the entire body of knowledge contained in its knowledge base while ensuring its consistency, notably by applying inference techniques. Inference allows the agent to derive new knowledge from existing knowledge; for example, inferring that a road is likely to be slippery if it is raining, or that a driver is likely to turn if they are signalling. However, inference requires a formal knowledge representation and management that current machine learning-based AI does not support.

### 3.3 Building the Predictive Model and Planning

The difficulty in developing predictive models and planning to achieve goals is the Achilles' heel of AI agents. Machine learning techniques are effective for perception, but they are ill-suited to reactive decision-making problems. Indeed, these problems require the construction and analysis of dynamic data with a high degree of precision, particularly when safety goals must be met. This is why AI agents are limited to planning in static environments where plans consist of sequences of actions.

The predictive model, based on a given state of the agent, is a transition system, a prediction tree, designed to represent the step-by-step interaction between the agent and its environment. The tree can be constructed from the agent's current state by alternately applying the agent's controllable actions and the uncontrollable actions performed by the environment. In other words, each state resulting from a controllable action is followed by states resulting from the execution of all applicable uncontrollable actions. Clearly, this theoretical construction implies exponential complexity in the space of accessible states. It is applied to simple systems or to systems where exploration of reachable states up to a limited horizon is sufficient [29].

Consider the example of a chess-playing system. To choose its next move, it must determine its opponent's possible moves over a certain horizon, as far into the future as possible. The complexity increases exponentially as the horizon grows longer. When the rules of interaction are well understood and formalized, as is the case here, it is possible to construct useful approximations of this model using Monte Carlo tree search (MCTS) algorithms, as in AlphaGo [30].

However, interactions between the agent and its environment are rarely ordered or synchronized. In current MAS, machine learning techniques are largely insufficient for reactive goal management and agent planning. Solutions using Chain of Thought (CoT) [31] or Tree of Thoughts (ToT) [32] generate plans that, in their final executable form, reduce to linear sequences of actions. While CoT produces a single chain, ToT improves upon this by searching over multiple branching reasoning paths and enabling backtracking [32]. However, this search occurs entirely within a static, internal problem space; ToT does not interact with or receive feedback from a dynamically changing environment during plan generation. Consequently, while these planners are sufficient for offline tasks, such as planning a trip or scheduling a meeting, they are incapable of generating plans that require real-time reactivity to exogenous events or state changes in a live MAS environment.

Consider the example of a real estate agent who, faced with fluctuating real estate prices, must anticipate changes in the behavior of their partners and identify the steps necessary to close a deal. This requires planning over time, anticipating the reactions of other agents, and adapting to changing conditions. Current AI planners cannot handle this level of complexity because they treat planning as a sequence of actions rather than as a dynamic interaction with other agents. The same applies to an autonomous vehicle or an energy management and distribution system within a smart grid.

We need planners capable of generating subtrees of the predictive model such that, given uncontrollable environmental actions, the execution of controllable actions preserves the validity of the invariant goals while maintaining the possibility of achieving the transient goals.

Some advocate the use of **reinforcement learning** techniques or training the model using a large number of dynamic scenarios to generate plans. However, these techniques rely on optimization methods that can address the problem of achieving transient goals to a certain extent, but cannot account for invariant goals, which ensure the safety of actions. As established in the safe reinforcement learning literature [33], standard RL optimizes expected return and does not inherently guarantee that safety constraints are satisfied at all times. Even with extensive training on dynamic scenarios, agents may violate constraints in previously unobserved states [34]. In robotics applications, this limitation is particularly critical, as real-world deployment requires guarantees that unsafe actions are never taken [35].

More recently, solutions based on **world models** have also emerged [8]. As explained in 2.1, these models operate on a representation of the world learned through training on videos illustrating an agent's (e.g., a robot's) interactions with their environment. Unlike generative models that predict every pixel, world models learn in a compressed latent space, making them computationally efficient while capturing physical concepts such as gravity, momentum, and object permanence [36]. They can make predictions by simulating possible evolutions of the system starting from a given state. However, it is unlikely that they can capture the dynamic properties of objects in their environment through image analysis alone.

## 3.4 AI Meets Systems Engineering: Achieving Collective Intelligence

Here we address an issue of great importance to multi-agent systems (MAS): the organization of agents within a community that coordinates its efforts to achieve the system's overall goals and demonstrate collective intelligence.

### *3.4.1 Distributed Computation and Coordination*

MAS are fundamentally component-based systems, where each component is an agent. No matter how intelligent each agent may be individually, the effectiveness of the system as a whole will depend on agent coordination, which inevitably raises questions regarding systems engineering.

How to coordinate multiple agents into a coherent system? An agent is not merely a mathematical concept. Ultimately, it must be implemented in a digital environment where hardware and software are interconnected. The same multi-agent system can be implemented in different ways with regard to the coordination of its agents within an architecture.

Autonomous systems involving numerous agents spanning a large area will not be free of undesirable emergent properties such as deadlocks or livelocks, even if each agent considered separately is shown to be safe. A system composed of individual, harmless agents can give rise to dangerous behavior as a whole that is difficult to analyze and verify after the fact. This is why it is important to identify and prevent dangerous situations from arising as early as the design phase of the MAS.

The problem of choosing an architecture requires taking into account the requirements related to logical relationships in communication and institutional relationships among agents. A centralized architecture, with a single central agent managing all interactions, is impractical for large-scale systems due to scalability, fault tolerance, and latency concerns. However, complete decentralization also has drawbacks: coordination becomes more difficult, and emergent behavior is harder to predict. The optimal architecture likely lies somewhere between these extremes, with some hierarchical structure to manage complexity.

If a MAS is to have a structure similar to that of human societies, its architecture must satisfy the following properties:

**Temporal Dynamism.** Over time, agents may be created or disappear according to rules related to the agents' own conditions of existence, as well as to collective rules. For example, an agent may disconnect due to a malfunction that takes it offline, but may reappear if it manages to recover. However, one can also imagine an agent being expelled from its community through the application of collective rules. Consider a fleet of autonomous delivery robots: over time, robots join the fleet, break down, and are replaced.

**Spatial Dynamism.** The relationships between agents depend on their position in a physical or virtual space that can change over time. A mobile agent must be able to incorporate changes occurring within its network of communicating neighbors into its perception system. The robots move through physical space, forming temporary clusters to coordinate deliveries.

**Organizational Dynamism.** Relationships between agents may depend on their temporary or permanent membership in an organization. An agent can dynamically join a group of agents to perform a task; for example, to improve highway traffic efficiency, autonomous vehicles can form platoons that vehicles dynamically join and leave. This requires the implementation of a distributed algorithm in which the lead and rear vehicles in the platoon play specific roles.
Alternatively, an agent can join or leave an institution, which confers rights and obligations implemented through dynamic updates to its attributes. This is known as organizational dynamism, which is very important for achieving collective intelligence, as seen in all animal societies; for example, ants' stigmergy algorithms (where agents communicate indirectly through modifications to the environment).

These three types of dynamism correspond to changes in the agent's environment (World) and in the agent itself (Self). Temporal dynamism affects the composition of the system; spatial dynamism affects the interaction relationships between agents; organizational dynamism affects the roles and obligations of agents. All three must be addressed by the agent's knowledge base and operating conditions.

A more in-depth analysis of these issues is beyond the scope of this article. It should be noted, however, that the development of MAS could greatly benefit from the application of the theory of algorithms and distributed systems, particularly a significant body of results concerning protocols for reaching consensus or preserving the integrity of shared partial knowledge.
With MAS, the classic problems of distributed dynamic systems are compounded by additional challenges specific to AI agents, since the messages exchanged contain unstructured data, such as linguistic data. Rigorously solving these problems in the context of MAS will require adapting and extending existing methods. Such results are sorely lacking, judging by the coordination problems currently posed by the construction of MAS.

#### *3.4.2 Trust, Communication, and Institutional Organization*

The effective integration of agents into an autonomous system requires alignment between the agents' needs and roles in terms of communication and institutional relationship management: why agents communicate, what motivates them to share knowledge and cooperate, and what is the rationale for the application of institutional authority?

It is particularly worth noting that there are three areas of agency—physical, communicative, and institutional—corresponding to three distinct types of agents, each of which requires its own specific form of integration:

- The **Physical Agent** operates in a physical environment using sensors and actuators, through interfaces that are largely predefined and fixed.

- The **Communicator** operates in a communication world. Its infrastructure, which includes communication channels, protocols, message formats, must be specified for each agent in a MAS. In addition, the agent must have a thorough knowledge of its environment: who it can communicate with, what messages make sense, and what norms govern its interactions.

- The **Institutor** operates in an institutional world. Its infrastructure, including norms, roles, authority relationships, reporting structures, must also be specified for each agent. Agents need not only access to institutional mechanisms but also knowledge of their "institutional neighborhood": who has authority over them, who they have authority over, and what norms apply to their actions.

For every domain, the challenge is the same: technical integration must go hand in hand with functional and cognitive integration. Below, we address these issues as they relate to a Communicator and an Institutor.

**Communication**

What motivates an autonomous agent to communicate? This motivation must be linked to the agent's cognitive states; in particular, its awareness of gaps in its own knowledge and its ability to recognize that another agent might fill those gaps. For example, if it realizes it needs knowledge that another agent can provide, or if it needs a "helping hand" to act with others to accomplish a task, it will initiate communication.

But communication requires more than motivation. It requires two forms of integration:

- **Technical integration:** the agent must have access to communication channels and protocols that enable it to send and receive messages.

- **Cognitive integration:** the agent must understand the meaning of the messages it sends and receives, the intentions of other agents, and the norms that govern communication.

Consider the example of a delivery robot in a smart factory. To coordinate with other robots, it must:

- **Technically:** share communication channels, message formats, and protocols with other robots.

- **Cognitively:** understand what other robots are communicating, recognize when it needs help, and interpret the intentions of its peers.

If the robot has technical integration but not cognitive integration, it can send and receive messages but cannot understand them. If it has cognitive integration but not technical integration, it understands what it should communicate but cannot do so.

In both cases, the question of trust arises. Should I rely on the information provided by another agent or accept another agent's collaboration to accomplish a task together? Trust is particularly important for communicators: communication is the primary medium through which trust is established, maintained, and repaired. Trust is the foundation of all social interaction. In MAS, trust must be established through mechanisms that are not yet well understood. Each agent must establish, based on observations or data regarding the agents' identities and capabilities, trust indicators as well as rules for using these indicators.

A more fundamental question is: Why would an agent communicate at all? If each agent has its own goals and priorities, why would it share knowledge or cooperate? This is the classic "tragedy of the commons": in the absence of mechanisms to align individual incentives with collective goals, agents may behave selfishly: by hoarding information, taking advantage of others' efforts, or pursuing their own goals at the expense of the system. To prevent this, the autonomous system must have mechanisms, such as norms, incentives, and sanctions that align individual and collective interests. This is not a technical problem that can be solved with better protocols; it is a social problem that requires norms, incentives, and governance structures.

**Institutional Organization**

Just as communication requires integration across technical and cognitive dimensions, so too does institutional organization.

The effective achievement of the overall goals of a system of autonomous agents depends on the establishment of authority relationships among agents. Each agent applies institutional norms individually. It knows what is permitted, required, or prohibited to do based on its role and the applicable rules. But the central challenge is this: how does a coherent global institutional organization emerge from the individual application of norms by multiple agents?

Like communication, institutional action requires two forms of integration:

- **Technical integration:** the agent must have access to the institutional mechanisms—databases, reporting structures, authorization systems—that enable it to exercise its authority.
- **Institutional integration:** the agent must understand its role, the norms that govern its actions, and its place in the hierarchy of authority.

Consider the example of a police agent. To function effectively, it must be integrated into the police power system both technically and institutionally:

- **Technically:** it must share common resources (databases, communication channels, sensor networks) and be able to exchange information with other agents in the system.
- **Institutionally:** it must report to the hierarchy, receive orders from authorized superiors, and exercise its authority within the limits defined by the applicable norms.

These two dimensions must work together. Technical integration without institutional integration leads to chaos: agents can communicate but have no shared understanding of their roles and obligations. Institutional integration without technical integration leads to ineffectiveness: agents understand their roles but cannot coordinate their actions. An agent may know that it has the authority to make an arrest, but if it cannot communicate this to other agents, its authority is meaningless.

Trust is also relevant to institutional organization, though perhaps less central than in communication. Institutional authority derives its legitimacy from the trust that the system is just, that rules are enforced fairly, and that authority is exercised appropriately. But for an artificial agent, institutional trust is primarily about reliability and predictability: can I trust that the system will enforce the rules consistently? Can I trust that my superiors will issue legitimate orders?

The choice of power architecture is essential to ensuring the effective exercise of institutional authority. A hierarchical structure, where authority flows from top to bottom, is efficient for command and control, but it is vulnerable to failures at higher levels. A distributed structure, where authority is shared among multiple agents, is more resilient, but it can be inefficient and may lead to conflicts. The optimal architecture depends on the specific requirements of the system: the number of agents, the criticality of the tasks, and the need for resilience versus efficiency.

In all cases, the key lesson from institutional theory in the social and political sciences is that institutional organization cannot be reduced to technical infrastructure. It requires the alignment of norms, roles, authority, and the coordination mechanisms that support them. This alignment is precisely what the knowledge attributes framework, such as mission, faculties, integration conditions, and operating conditions, is designed to capture.

# 4. The Two Faces of Trustworthiness: Behavioral and Cognitive Properties

## 4.1 Behavioral vs. Cognitive Properties: Why Behavior Is Not Enough

For AI systems, trustworthiness cannot be limited to behavioral properties. These systems are designed to mimic human cognitive functions, in ways that resemble human intelligence. This introduces a new category of properties, cognitive properties that depend not only on what the system does, but also on the way in which it decides to do it.

### 4.1.1 What the Agent Does: Behavioral Properties

Behavioral properties are properties that can be defined based on a system's observed input-output behavior, independently of its other structural and physical properties. There are three distinct types of behavioral properties.

**Safety properties** guarantee that the system will never enter dangerous states for different reasons, including flaws in the fabrication process or damages caused through interaction with its environment, including its authorized users. They identify conditions on the agent state that should be avoided at the risk of damaging the system itself or its environment.

**Security properties** guarantee that the agent will not be accessed by anyone other than its authorized users. Agents should have predefined rules that impose access control on their various users. Security issues arise when unauthorized users gain access to the agent either by impersonating a normal user or by surreptitiously penetrating the agent's access control. Security properties can be refined to cover aspects such as confidentiality, integrity, and availability.

**Usefulness properties** characterize the extent to which an agent effectively meets user needs and provides value by achieving its intended goals. These include properties such as functionality (does the agent do what it is supposed to do?), performance (does it do it fast enough?), efficiency (does it do it with reasonable resources?), and usability (can users interact with it effectively?).

All the essential properties of traditional systems are behavioral properties. These properties can be defined in a way that is relevant to AI agents, despite the inherent difficulties in formalizing them. However, for AI agents designed to mimic human cognitive processes, behavioral properties do not capture the full extent of what makes them trustworthy. We need a new category of properties: cognitive properties.

| Agent Properties | Behavioral Properties | | Cognitive Properties |
|---|---|---|---|
| Requirements | | | |
| Risk-related properties (no negative outcomes) | Safety | Security | Prohibitive properties (ethical/law-enforced) |
| Progress properties (positive outcomes) | Functionality, Performance, Efficiency, User-friendliness | | Performative properties (Intent and Goal-directed) |

Figure 3: Behavioral vs. Cognitive properties.

Figure 3 shows the classification of the different types of behavioral and cognitive properties following the standard distinction between risk-related and progress properties.

### 4.1.2 How the Agent Thinks: Cognitive Properties

Cognitive properties are human-centric properties that cannot be determined from observed external behavior. Unlike behavioral properties, which can be verified by observing inputs and outputs, cognitive properties require access to the agent's internal knowledge, reasoning processes, and value systems. They are properties of how the agent thinks, not just what it does.

To understand these properties, one must take into account the fact that agent choices and decisions are determined by norms and value systems as explained in Section 2.2. Norms define which actions are permitted, required, or prohibited; for example: "I must not lie" or "I must not run a red light." In addition, value systems are used to evaluate the costs and benefits arising from actions taken for oneself and for one's environment; for example, if I lie, my credibility is at stake; if I run a red light, my driver's license will be suspended.

It is easy to see how an agent's knowledge and the application of norms affect the validity of cognitive properties. If an agent asserts, "The Earth is flat," we cannot determine whether it is lying or simply unaware of this fact without knowing the exact extent of its knowledge. To determine that the agent is lying, we would need to know that it is equipped with norms that forbid knowingly asserting a false proposition.
Similarly, if an agent chooses to run a red light, we must determine whether it is out of ignorance of traffic rules or because it prefers to take that risk due to other priorities, knowing that this is forbidden.

Cognitive properties are of two types.

**Normative properties** are risk-related cognitive properties that characterize the conformity of agent decisions with norms such as legal or moral rules. They express the fact that the agent does not violate rules of its own normative system.

**Intent-directed properties** characterize the way the agent decides by choosing and planning its goals. Their satisfaction depends on the agent's knowledge and reasoning capabilities.

The distinction between behavioral and cognitive properties has profound implications for how we evaluate and trust AI systems. It is emphasized by the Chinese Room argument [37], which shows that behavioral tests cannot distinguish between a simulation of cognitive functions and their actual execution based on understanding. It also reflects two different approaches in psychology: Behaviorism, which seeks to understand behavior by measuring only observable conditions and events; and Cognitivism, which emphasizes the role of knowledge and reasoning as determinants of behavior.

Together, normative and intent-directed properties can be used to characterize two essential dimensions of rational agency: compliance (following the rules) and deliberation (making good choices). A trustworthy agent must do both: it must respect the norms that govern its behavior, and it must make decisions that are optimal, consistent, and competent.

**Rationality** is a property specific to human deliberative processes. It encompasses various intent-related properties that play an important role in assessing an agent's trustworthiness:

1. Optimal decision-making and choosing among permissible goals or actions, guided by their utility as determined by value systems.
2. Consistency of solutions: problems posed in logically equivalent situations give rise to similar solutions.
3. Competence levels: passing a test at one level implies passing a test at a lower competence level.

It should be noted that if an agent is rational, understanding, analyzing, and validating its behavioral properties can be greatly simplified. Many testing methods for autonomous driving agents assume rationality when selecting test cases. For example, they focus on simple configurations, assuming that for more complex but logically equivalent configurations, the test results will be similar. Take, for example, the test scenario in which a car is driving in a lane, following a vehicle ahead at a safe distance. It can be inferred that if, in this scenario, other vehicles traveling in other lanes are added, this will not affect the test result.
Furthermore, assuming that self-driving agents act rationally, often test scenarios focus on the most critical situations. if a vehicle can successfully navigate a critical scenario, it will then be able to successfully handle a less critical scenario—achieved, for example, by increasing the distances between the vehicles involved or simply by removing certain vehicles from the scenario. It should be emphasized that experimental results of simulated scenarios show that AI driving agents are not rational [38]. This is not surprising: rationality requires a level of understanding and consistency that current AI systems, with their empirical knowledge and lack of common sense, cannot provide.

### 4.1.3 Why Cognitive Properties Matter: Two Examples

Consider an AI driving system that passes the standard tests given to human drivers. Would this be sufficient to authorize it to drive? The obvious answer is no. But analyzing why this test is insufficient leads us to consider the cognitive properties we assume for humans but cannot guarantee for artificial agents.

A human driver who passes the driving test is assumed to possess:

- A mental model of the world;
- An understanding of traffic rules;
- Awareness of other drivers' intentions;

- The capacity to make ethical judgments in unforeseen circumstances.

An AI that passes the same behavioral test, by contrast, possesses none of these. It has simply demonstrated that its outputs can, under controlled conditions, mimic those of a competent driver. The difference between behavioral mimicry and true cognitive reliability is not a matter of philosophical abstraction. It is the difference between a system capable of functioning under specific conditions and a system that can be relied upon in any context; a system that takes into account not only technical rules but also legal and ethical standards in the pursuit of its goals. The assessment of cognitive properties in ML-based AI systems is particularly challenging because they have very limited reasoning capabilities and lack a conceptual model of the world equipped with a value system capable of guiding their decision-making processes.

The driving test example reveals what AI systems lack. But the problem extends beyond driving. Consider a more profound case: the medical exam.

There is every reason to believe that an AI agent could pass final medical exams just as well as human students. What does this imply? That the AI agent should be allowed to practice medicine? The answer is no, not because the AI lacks medical knowledge, but because it lacks the cognitive properties that make a physician trustworthy: empathy, judgment, accountability, and the ability to explain and justify decisions. These are not properties that can be tested by multiple-choice questions.

In addition to achieving a certain level of expertise attested by appropriate benchmarks, we should have solid evidence that the system is ethically and rationally aligned with human values. This practically means that the system should satisfy additional cognitive properties that we consider granted for humans. In fact, we trust humans because we know that they are aware of the rules of a common value system and are bound by them. They are liable for their acts and may be punished if they fail to respect the rules. However, it does not make sense to put a machine in jail.

We often see studies that seek to demonstrate that machines are more intelligent than humans using behavioral tests in the form of question-and-answer sessions. In particular, we see statements from tech giants claiming that their systems outperform experts in diagnostic analysis, alongside a multitude of articles announcing that a particular system has passed the Turing test or achieved benchmark scores proving its superintelligence. These comparisons are superficial and are intended solely for marketing and sales promotion purposes.

## 4.2 The Responsible AI Mirage: A Critical Assessment

Despite the numerous studies on "trustworthy AI," there is currently considerable confusion regarding its meaning for AI systems and the methods for guaranteeing its characteristic properties. In particular, the study of cognitive properties has given rise to a growing body of literature on "responsible AI," "aligned AI," and "ethical AI" [5]. Furthermore, these terms feature prominently in the marketing campaigns of major technology companies, which present them as essential attributes of their products.

Unfortunately, all this talk about cognitive properties seems to miss their true nature. Current studies evaluate cognitive properties based on behavioral criteria [6]. However, responsibility requires more than just behaving in a certain way. It requires that the agent acts with intent, that it can be held accountable for its actions, that it understands and respects the norms that govern its choices, and that it can bear the consequences of its actions. These conditions are met by human agents, but not by current AI systems.

Therefore, these properties can only be attributed to agents whose architecture makes their knowledge accessible, as in the proposed agent model architecture, where the knowledge base contains the agent's goals, norms, and value systems, and the reactive and proactive modules deliberate about actions. This is precisely what current AI systems lack.

It is remarkable that public and private institutions are publishing guidelines, frameworks, and principles setting out requirements regarding the cognitive properties of AI. However, all of this remains wishful thinking and is impossible to implement in practice. Some research seeks to distil these into a few fundamental principles [39]. Their work, and others like it, identifies principles such as fairness, accountability, and transparency. These are admirable goals, and such frameworks serve an important role in shaping regulatory discourse. However, these works do not provide methods for implementing them [40]. All these ideas sound wonderful. However, when it comes to putting them into practice, problems arise: fairness is contested, accountability lacks clear metrics, and transparency often reduces to superficial disclosures.

Not only do such frameworks often lack concrete technical guidance, but they also add to the confusion about what can reasonably be expected from AI systems and their developers. This is particularly true of many articles on the ethics of AI, which present as ethical problems what are in reality merely violations of safety rules.The fact that a Tesla driver was killed in a traffic accident after Autopilot failed to recognize an oncoming truck clearly raises a safety issue, not an ethical one [41] [42]. This distinction is crucial: safety violations are failures of the system to meet its technical requirements; ethical violations require deliberate choices that violate moral norms. Conflating the two confuses the public and obscures the real problems. The Tesla accident is the result of a system failure that has nothing to do with making an ethical choice.
Similarly, racist or sexist remarks made by a chatbot, or instructions on how to make a bomb, reflect the synthesis of information contained in its training dataset [43]. These are violations of safety properties that have nothing to do with a deliberate choice to transgress ethical rules, as long as we refer solely to the observed behavior. Following this logic, a process scheduler could be considered unethical if the way it distributes processes among processors is deemed unfair. This absurd conclusion reveals the flaw in the reasoning: fairness in a process scheduler is a technical property (e.g., ensuring no process is starved), not an ethical one. Conflating technical fairness with ethical fairness is a category mistake that only adds to the confusion.

Other examples cited in the articles on ethical AI concern anthropogenic risks that clearly involve human responsibility. Why does a privacy breach, which in traditional systems is viewed as a security issue, suddenly become a violation of an ethical rule by the system itself? The answer is that it does not. A privacy breach is a security failure. It may have ethical implications, but the breach itself is a technical failure. The system is not making an ethical choice; it is failing to enforce its access controls.

If criminals use AI technology to harm others or society, they are clearly liable under the law. Impersonating a corporate executive and requesting a fraudulent wire transfer constitutes a reprehensible act. Of course, the security of the system used may also be called into question, as it failed to protect the access rights to the system used for this purpose.

In conclusion, current research and debates on responsible and ethical AI miss the point, because validating cognitive properties requires access to the system's internal knowledge and its reasoning processes. This seems extremely difficult, if not impossible, given the current state of the art.

## 4.3 Specifying and Validating Trustworthiness

We examine the problem of specifying and validating the trustworthiness of AI systems, which has two dimensions, corresponding to the two types of properties we have identified. With regard to behavioral properties, the challenge lies in validation: can we test AI systems with the same degree of confidence as traditional systems? As for cognitive properties, the challenge is even more fundamental: the question remains how to define them rigorously for current AI systems, let alone validate them.

### 4.3.1 Testing Behavioral Properties

Due to their lack of explainability, the properties of neural networks can only be validated using empirical techniques, that is, through testing. This significantly limits the degree of confidence in the validation results, which is considerably lower than that obtained by verifying traditional systems which can be modelled as transition systems. Thus, behavioral properties of AI can only be refuted, as testing can reveal the presence of defects but cannot prove their absence. In traditional systems, verification can provide this proof; in AI systems, it cannot.

A fundamental challenge is the question of coverage. For traditional systems, we have coverage criteria that involve rules ensuring that we have exercised all the important parts of the system. We can test every line of code or every logical branch because traditional systems are built from well-defined components that we can understand and analyze. For AI systems, we have no such criteria. The system is a black box. We cannot analyze its internal structure in a meaningful way.

A testing method that circumvents the difficulty posed by the lack of a structure for calculating test coverage involves validation through functional test scenarios targeting a specific expected property. This type of functional testing is often applied to complex systems that interact with an environment. A key criterion for selecting test cases is to consider worst-case scenarios in which the system must handle situations deemed "difficult" according to a given criterion. Thus, for real-time systems, tests are conducted to evaluate worst-case execution times or to assess performance under maximum load conditions for a given task. This approach is based on the assumption that "if it can handle the worst, it can handle the best," which is consistent with the assumption that the system behavior is rational. As explained in Section 4.1.2, experimental results show that AI agents do not possess this property. An autonomous agent capable of navigating a complex intersection may still fail in a simpler scenario. This fact directly raises the question of the value of worst-case testing methods for AI agents and further complicates the problem of testing them.

An important criterion regarding the validity of test results is the distinction between technical systems and non-technical systems.

A **technical system** is a system whose observed behavior can be unambiguously characterized as a relationship between the mathematical domains of its inputs and outputs. In other words, for a given input, the correct output is defined by a predicate that characterizes the tested input-output relationship.

Traditional ICT systems, whether hardware or software, are technical systems, as are a minority of AI systems, such as chess-playing systems or an autopilot, which involve numerical inputs and outputs.

A **non-technical system**, on the other hand, involves sensory or linguistic inputs and outputs and therefore requires interpretation. Large language models (LLMs) are the most characteristic example of this. Their behavior does not lend itself to formalization, and their properties cannot be defined objectively, since the relationship between inputs and outputs is subject to interpretation. To determine that the correct answer to the question "What is the capital of France?" is "Paris," one must first interpret this correspondence. There is no formal rule for deducing "Paris" from the question: only a statistical correlation and human judgment.

This distinction, which is rarely emphasized in the literature, has obvious implications for the objectivity of test results.

#### *4.3.1.1 Technical Systems: Testing Autonomous Vehicles*

To illustrate the challenges encountered during the validation of technical systems, we will use a representative example: autonomous vehicles. As explained above, it is not possible to define simple and relevant coverage criteria for these systems, as is the case with software and hardware. Very often, manufacturers cite the number of miles travelled in autonomous mode during simulations as a criterion. It is clear that this criterion has no technical value, since one must explain how the simulated miles correspond to "real" miles. We need evidence, based on coverage criteria, demonstrating that the simulation does indeed account for the many different situations, such as various road types, traffic conditions, weather conditions, and so on. Without such criteria, it is impossible to answer a crucial question: at what point is a tested vehicle sufficiently safe?

For autonomous vehicles, there is a considerable, if not infinite, number of scenarios. Consequently, random testing cannot provide sufficient evidence, even with a very large number of simulated miles. We need criteria for selecting scenarios in order to cover, with a high degree of probability, all hazardous situations. These criteria should be based on a statistical analysis of real-world situations characterized by complex and semantically rich data regarding the physical environment, traffic signs, and the position of obstacles near the test vehicle.

For example, with regard to autonomous vehicles, the application of traditional testing methods raises the following two issues.

First, tests must cover the wide variety of static road configurations determined by purely topological relationships. For example, a roundabout, a highway on-ramp, or an intersection where access is controlled by "Stop" signs or traffic lights.

Second, for each static configuration, tests must cover the wide variety of dynamic scenarios involving the kinematic state of the vehicle under test and that of the surrounding obstacles.

We are therefore faced with a two-dimensional complexity: we must cover both the diversity of static configurations (roundabouts, intersections, highways) and that of dynamic scenarios (variable number of obstacles, different speeds, variable weather conditions). The number of combinations is enormous, and we have no coverage criteria to guide the selection of tests. As already explained for dynamic configurations, success in the most difficult situations does not guarantee success in easier situations obtained simply by reducing the number of obstacles surrounding the tested vehicle or by relaxing the dynamic constraints [38].

A direct consequence of the above is that obtaining robust guarantees regarding the behavioral properties of technical AI systems remains a challenge, at least from the perspective of the current state of knowledge. And we can already predict that we will not be able to achieve the level of assurance obtained for traditional critical systems through the use of mathematical models implemented by a wide range of methods, ranging from verification and static analysis to white-box testing and stochastic analysis techniques. The only way to compensate for these insufficient safety guarantees might lie in the use of explicit knowledge from an agent built according to the reference architecture. However, this is merely a potential avenue, not a proven solution.

#### *4.3.1.2 Non-Technical Systems: The Interpretation Problem*

For non-technical systems such as LLMs, test results are necessarily subjective, as the validity of the relationship being tested depends on human interpretation. In some cases, when the relationship refers to facts, for example, what is the capital of a country, it can be validated unambiguously. However, if the property being tested concerns the absence of bias or the fairness of the response, the evaluation results have no objective value.

Another problem that may arise, particularly for LLMs, is the lack of coverage criteria to guide the selection of tests to be applied. Without delving too deeply into this topic, it seems difficult to imagine a coverage function for natural language queries. We are therefore led to rely on benchmarks developed by experts, which include test cases and acceptable margins of error in the interpretation of responses by human "oracles".

A detailed analysis of the risks associated with the use of LLMs and the corresponding evaluation methods is presented in [44]. The authors argue that the stochastic nature of LLMs, combined with their lack of interpretability, makes them fundamentally unsuitable for applications where reliability is essential. Their analysis reinforces the concerns we have raised here. It is important, however, to understand these limitations of LLMs and to put their success into context, given their role in building intelligent systems, particularly autonomous systems where reliability is paramount.

### 4.3.2 Cognitive Properties: The Ontological Gap

Recent developments in AI agents propose architectures that integrate essential autonomic functions such as perception, goal management, planning, and memory. These architectures could serve as a basis for testing cognitive properties if they allow access to an agent's explicit knowledge and intentions. In such an architecture, cognitive properties could, in principle, be specified using agent logics and verified by examining the knowledge base and the reasoning processes of the reactive and proactive modules. There is a substantial body of literature on agent logics, such as epistemic and deontic logics, for expressing these properties [26][45][46][47]. Using notations proposed by these logics we can express agent cognitive properties.

For instance, an agent *x* is dishonest if it satisfies the formula: *says(x, p) ∧ knows(x, ¬p)*, meaning that it asserts *p* while *x* knows *p* is false. The evaluation of *knows(x, ¬p)* requires access to the agent's knowledge base.

Similarly, agent *x* does not act responsibly if *x* performs action *a* knowing that *a* is prohibited according to the norms $N$ stored in its knowledge base:
$does(x, a) \land knows(x, \neg right(x, do(x, a), N))$.
The predicate $right(x, do(x, a), N)$ is true only if the norms $N$ permit action *a*.

Finally, agent *x* is not held responsible for committing a prohibited act *a* if:
$does(x, a) \land \neg knows(x, \neg right(x, do(x, a), N))$.

However, beyond the issue of formalization, a fundamental question remains unanswered: what does it mean to attribute cognitive properties to an artificial agent? For example, acting responsibly implies that the agent acts intentionally, can be held accountable, understands the norms, and is capable of bearing the consequences of its actions. If an artificial agent does not meet these conditions, it cannot be considered responsible.

Let us imagine that we want to create an ethical agent that conforms to human ethical values, as specified by norms that the agent can use to make decisions. While humans may knowingly violate ethical rules due to various intrinsic biological needs, the "needs" of AI agents will be defined by their developers. Human values are rooted in empirical knowledge and experience; AI norms and value systems are specified by developers and implemented as constraints. This ontological difference is fundamental.
We can develop AI agents equipped with a normative system that prevents them from running a red light, lying, or violating fairness rules. Failure to adhere to these principles constitutes a safety breach. If they are designed to override these norms in specific cases, this can also be predicted and verified. For humans, ethical considerations may pertain to human well-being, or even life itself, if vital goals are not met. In contrast, the purpose of machines is, in principle, determined by their developers or their human users.

On the other hand, human agents enjoy a high degree of freedom, even when it comes to malicious or illegal actions. However, if they deliberately commit an offense, they are held accountable, which can result in penalties with economic consequences or deprivation of liberty. We cannot grant machines the same degree of freedom. If a machine decides to kill someone, even if we have assigned extremely high costs to that act, the penalty does not have the same effect as it does on humans. For human beings, the deprivation of material resources, freedom, or recognition can be a source of suffering and pain, or even a threat to their survival. For machines, punishment would amount to negative scores in a fictional value system, but one cannot cause a machine to suffer in the same way that one causes a human being to suffer by punishing them.
This is why trust in machines is fundamentally different from trust in humans. Trust in human beings is based on existential conditions, the violation of which does not have the same consequences for machines. This ontological difference between human beings and machines raises the question of to what extent it is appropriate to seek to align machines with human values.

Of course, we can anthropomorphize the behavior of agents by presenting them with dilemmas and evaluating it against ethical rules, which, incidentally, is at the heart of much of the research. But as we have shown, this amounts to adding an interpretive layer to observed behavior. For example, it is not meaningful to evaluate the ethical behavior of an autonomous car in "trolley problem" scenarios [48] based solely on observing their outcomes, since decisions may be purely random and the autopilot may not even be aware of possible alternatives. The car may appear to make an ethical choice, but it may be doing so without any understanding of the choice it is making, as illustrated by the Chinese room argument [37].

## 5 From Vision to Reality: The Path Forward

The challenges discussed reveal a fundamental truth: building autonomous systems is not merely a matter of scaling up existing techniques. It requires solving deep problems in perception, knowledge representation, decision-making, and systems integration.

The data from real-world deployments confirms this gap. Gartner predicts that more than 40% of agentic AI projects will be cancelled by the end of 2027 due to escalating costs, unclear business value, and inadequate risk controls [49]. Most projects are still early-stage experiments or proofs of concept, driven more by hype than by operational reality, and many vendors have rebranded existing chatbots as "agents" without delivering meaningful outcomes, a practice known as "agent washing" [50][51]. Gartner estimates that of the thousands of companies claiming agentic capabilities, only about 130 are building anything that genuinely deserves the label [49].

The gap between the proof of concept and production is particularly stark. Forrester's 2026 assessment found roughly three-quarters of enterprises adopting agentic AI, but only a tiny minority running it in real production [51]. The problem is a "capability-deployment verification gap": the agent can perform the task in a controlled test, but the business cannot verify or trust it once it runs against proprietary systems and live data [49].

Industry is investing heavily in MAS without having solved the fundamental problems of reliability, communication, and institutional organization. The protocols being developed provide the plumbing, the channels through which agents can communicate, but they do not solve the deeper problems of what agents should communicate, why they would choose to do so, or how to ensure that their interactions are reliable and trustworthy.

Proponents of artificial general intelligence (AGI) suggest that AI has already achieved its goals while remaining strictly confined to the limits of machine learning. Our analysis strongly contradicts this view and shows that there is still a long way to go before we can cover a wide range of potential applications. Instead of giving up and clinging to the illusion that scaling up models will give us smarter agents, we propose a fusion of machine learning and symbolic AI, all within a technical framework that updates and extends traditional systems engineering, because, ultimately, AI systems are built from hardware and software.

The main result of this study is the agent’s reference architecture, which goes beyond simple, specific solutions and highlights the diversity of cognitive behavior resulting from the choice of the agent’s knowledge attributes. We have demonstrated, to the extent possible, the generality of our architecture, which can be specialized for different types of missions and environments. Implementing agents based on this architecture poses challenges.

The systems engineering complexity of autonomous systems places them among the most difficult systems ever built. The combination of reactive complexity and architectural complexity, coupled with the dynamism and distributed nature of multi-agent systems, makes formal validation impossible.

Autonomous systems will not be free of undesirable emergent properties, even if each agent considered separately is shown to be safe.

One of the main challenges in designing autonomous artificial systems is fostering synergy and cooperation among agents. The difficulties encountered during the deployment of robotaxis in American cities are a prime example of this problem. A vehicle might prioritize the efficiency of its own route, thereby contributing to traffic jams. Or it might block an ambulance or a police car because its programming fails to recognize the need to yield to emergency vehicles [52]. In one specific case, a fleet of autonomous vehicles blocked an intersection for hours because the optimization algorithm of each vehicle led to a total traffic jam [53], a situation that no single vehicle could resolve on its own. Each vehicle was individually “right” to follow its programmed logic, but the collective result was a failure.

Evaluating AI trustworthiness presents two distinct challenges.
Behavioral properties can only be tested empirically, but extending existing testing methods is problematic because coverage criteria and “worst-case” techniques do not apply to AI systems. We must recognize the limits of behavioral testing. Passing a benchmark is not the same as being trustworthy. It simply means that the agent has demonstrated its skills under specific, controlled conditions.

Cognitive properties require access to internal knowledge and reasoning processes, which current AI systems do not provide. We must develop methods for specifying and validating cognitive properties. Agent logics, which extend modal logic with epistemic and deontic modalities, provide a formal framework for specifying properties such as rationality, consistency, and normative compliance. But formal specification is only the first step. We need methods for validating these properties in practice.

Beyond these technical challenges, there is a deeper ontological gap: machines cannot be held responsible in the way humans can, because they cannot suffer the consequences of punishment or feel remorse.

The problems mentioned should stimulate research efforts rather than lead us to give in to the temptation to lower our standards. The argument that AI systems are "good enough" for certain applications is a dangerous lure. If we cannot verify that a system satisfies the cognitive properties required for trust, we should not deploy it in contexts where trust is essential.
We must be clear: a system that cannot explain its decisions is not transparent. A system that cannot be held accountable for its actions is not responsible. A system that does not understand the values it is supposed to uphold is not ethical. These are not properties that can be added to an opaque system; they are properties that must be built in from the start.

Until these obstacles are overcome, the potential of autonomous systems will remain largely untapped. The gap between vision and reality should spur us to take action to resolve these fundamental problems. The bet on multi-agent systems could still pay off, but only if we honestly assess our chances of success and commit to solving the fundamental problems we have identified.

## References

[1] D. Harel, A. Marron, and J. Sifakis, "Autonomics: In search of a foundation for next-generation autonomous systems," *PNAS*, vol. 117, no. 30, pp. 17491–17498, Jul. 2020. doi: 10.1073/pnas.2003162117.

[2] J. Sifakis and D. Harel, "Trustworthy Autonomous System Development," *ACM Transactions on Embedded Computing Systems*, vol. 22, no. 3, Article 40, pp. 1–24. doi: 10.1145/3545178.

[3] CBS News, "Elon Musk predicts Tesla will have 'robotaxis' on the road next year," Apr. 22, 2019. [Online]. Available: https://www.cbsnews.com/news/elon-musk-claims-tesla-will-have-self-driving-robotaxis-on-the-road-next-year/

[4] C. Huang, Z. Zhang, B. Mao, and X. Yao, "An Overview of Artificial Intelligence Ethics," *IEEE Transactions on Artificial Intelligence*, vol. 4, no. 4, pp. 799–819, Aug. 2023. doi: 10.1109/TAI.2022.3194503.

[5] A. S. Dhaigude and G. B. Kamath, "Mapping responsible artificial intelligence in business and management: Trends, influence, and emerging research directions," *ScienceDirect*, 2025. doi: 10.1016/j.techfore.2025.123890.

[6] F. Momentè et al., "Triangulating LLM Progress through Benchmarks, Games, and Cognitive Tests," *arXiv:2502.14359*, 2025.

[7] "Scaling gamble," personal communication.

[8] Y. LeCun, "A Path Towards Autonomous Machine Intelligence," OpenReview.net, 2022.

[9] M. Georgeff, B. Pell, M. Pollack, M. Tambe, and M. Wooldridge, "The Belief-Desire-Intention model of agency," in *Intelligent Agents V (ATAL'98)*, Lecture Notes in Computer Science, vol. 1555, pp. 1–10, Springer, 1999.

[10] M. E. Bratman, *Intention, Plans, and Practical Reason*. Harvard University Press, 1987.

[11] A. S. Rao, "AgentSpeak(L): BDI agents speak out in a logical computable language," in *European Workshop on Modelling Autonomous Agents in a Multi-Agent World*, Springer, pp. 42–55, 1996.

[12] M. Dastani, M. B. van Riemsdijk, and J.-J. C. Meyer, "Programming multi-agent systems," in *Handbook of Agent-Oriented Programming*, Springer, 2009.

[13] A. Sheth, K. Roy, and M. Gaur, "Neurosymbolic AI: Why, What, and How," *arXiv:2305.00813*, 2023.

[14] S. Hao et al., "Reasoning with Language Model is Planning with World Model," in *Proceedings of EMNLP 2023*, 2023. [Online]. Available: https://arxiv.org/abs/2305.14992

[15] IBM Corporation, "An Architectural Blueprint for Autonomic Computing," White Paper, 2006. [Online]. Available: https://www-03.ibm.com/autonomic/pdfs/AC%20Blueprint%20White%20Paper%20V7.pdf

[16] S. Kounev, J. O. Kephart, A. Milenkoski, and X. Zhu (Eds.), *Self-Aware Computing Systems*. Springer, 2017. doi: 10.1007/978-3-319-47474-8.

[17] O. Shehory and A. Sturm (Eds.), *Agent-Oriented Software Engineering: Reflections on Architectures, Methodologies, Languages, and Frameworks*. Springer, 2014. doi: 10.1007/978-3-642-54432-3.

[18] H. Du, S. Thudumu, R. Vasa, and K. Mouzakis, "A Survey on Context-Aware Multi-Agent Systems: Techniques, Challenges and Future Directions," *arXiv:2402.01968*, 2024.

[19] J. McCarthy and P. J. Hayes, "Some Philosophical Problems from the Standpoint of Artificial Intelligence," in *Machine Intelligence 4*, B. Meltzer and D. Michie, Eds., Edinburgh University Press, pp. 463–502, 1969.

[20] P. Lewis et al., "Retrieval-Augmented Generation for Knowledge-Intensive NLP Tasks," in *Advances in Neural Information Processing Systems (NeurIPS 2020)*, 2020. [Online]. Available: https://arxiv.org/abs/2005.11401

[21] V. Karpukhin et al., "Dense Passage Retrieval for Open-Domain Question Answering," in *Proceedings of EMNLP 2020*, pp. 6769–6781, 2020.

[22] H. Zamani et al., "Retrieval-Enhanced Machine Learning: A Review," in *Proceedings of SIGIR '22*, 2022.

[23] S. Pan et al., "Unifying Large Language Models and Knowledge Graphs: A Roadmap," *IEEE Transactions on Knowledge and Data Engineering*, vol. 36, no. 7, pp. 3246–3266, 2024. doi: 10.1109/TKDE.2024.3352100.

[24] D. Edge, H. Trinh, and N. Cheng, "From Local to Global: A Graph RAG Approach to Query-Focused Summarization," *arXiv:2404.16130*, 2024.

[25] G. Mialon et al., "Augmented Language Models: A Survey," *arXiv:2302.07842*, 2023.

[26] R. Fagin, J. Y. Halpern, Y. Moses, and M. Y. Vardi, *Reasoning About Knowledge*. MIT Press, 1995.

[27] D. Lee, C. Park, J. Kim, and H. Park, "MCS-SQL: Leveraging Multiple Prompts and Multiple-Choice Selection For Text-to-SQL Generation," in *Proceedings of COLING 2025*, pp. 337–353, 2025.

[28] "LinkAlign: Scalable Schema Linking for Real-World Large-Scale Multi-Database Text-to-SQL," in *Proceedings of EMNLP 2025*, 2025.

[29] O. Maler, A. Pnueli, and J. Sifakis, "On the synthesis of discrete controllers for timed systems," in *Proceedings of STACS'95*, Lecture Notes in Computer Science, vol. 900, Springer, pp. 229–242, 1995.

[30] D. Silver et al., "Mastering the game of Go with deep neural networks and tree search," *Nature*, vol. 529, no. 7587, pp. 484–489, Jan. 2016. doi: 10.1038/nature16961.

[31] J. Wei et al., "Chain-of-Thought Prompting Elicits Reasoning in Large Language Models," in *Advances in Neural Information Processing Systems*, 2022. [Online]. Available: https://arxiv.org/abs/2201.11903

[32] S. Yao et al., "Tree of Thoughts: Deliberate Problem Solving with Large Language Models," in *Advances in Neural Information Processing Systems (NeurIPS 2024)*, vol. 36, 2024.

[33] J. García and F. Fernández, "A comprehensive survey on safe reinforcement learning," *Journal of Machine Learning Research*, vol. 16, pp. 1437–1480, 2015.

[34] D. Amodei et al., "Concrete problems in AI safety," *arXiv:1606.06565*, 2016.

[35] J. Kober, J. A. Bagnell, and J. Peters, "Reinforcement learning in robotics: A survey," *International Journal of Robotics Research*, vol. 32, no. 11, pp. 1238–1274, 2013.

[36] Y. Matsuo et al., "Deep learning, reinforcement learning, and world models," *Neural Networks*, vol. 152, pp. 267–275, 2022.

[37] J. R. Searle, "Minds, Brains, and Programs," *Behavioral and Brain Sciences*, vol. 3, no. 3, pp. 417–457, 1980. doi: 10.1017/S0140525X00005756.

[38] C. Li, J. Sifakis, R. Yan, and J. Zhang, "Rigorous Simulation-based Testing for Autonomous Driving Systems – Targeting the Achilles' Heel of Four Open Autopilots," *arXiv:2405.16914*, 2024.

[39] L. Floridi and J. Cowls, "A Unified Framework of Five Principles for AI in Society," in *Ethics, Governance, and Policies in Artificial Intelligence*, Springer, pp. 5–17, 2021. doi: 10.1007/978-3-030-81907-1_2.

[40] B. Mittelstadt, "Principles alone cannot guarantee ethical AI," *Nature Machine Intelligence*, vol. 1, no. 11, pp. 501–507, 2019.

[41] A. Jatavallabha, "Tesla's Autopilot: Ethics and Tragedy," *arXiv:2409.17380*, 2024.

[42] N. Paulo, L. Kirchmair, and Y. E. Bigman, "Impartiality Preferences in Sacrificial Moral Dilemmas Involving Autonomous Vehicles," *Analysis*, anaf041, 2025. doi: 10.1093/analys/anaf041.

[43] TechRepublic, "AI Agent Reportedly Deletes Company's Entire Database, Admits to Violating Guardrails," May 4, 2026. [Online]. Available: https://www.techrepublic.com/article/ai-agent-deletes-company-database-admits-violating-guardrails/

[44] E. M. Bender, T. Gebru, A. McMillan-Major, and S. Shmitchell, "On the Dangers of Stochastic Parrots: Can Language Models Be Too Big?," in *Proceedings of FAccT '21*, pp. 610–623, 2021. doi: 10.1145/3442188.3445922.

[45] J.-J. Ch. Meyer and J. Treur (Eds.), "Agent-based defeasible control in dynamic environments," in *Handbook of Defeasible Reasoning and Uncertainty Management Systems*, vol. 7, Kluwer Academic, 2002.

[46] M. Wooldridge, *Reasoning about Rational Agents*. MIT Press, 2000.

[47] J. D. Burge and A. C. Esterline, "Modeling societies of agents using modal logics," in *Proceedings of IEEE SoutheastCon 2000*, pp. 75–82, 2000. doi: 10.1109/SECON.2000.845429.

[48] D. Cecchini, S. Brantley, and V. Dubljević, "Moral judgment in realistic traffic scenarios: moving beyond the trolley paradigm for ethics of autonomous vehicles," *AI and Society*, vol. 40, no. 2, pp. 1037–1048, 2025.

[49] Gartner, "Why 40% Of Agentic AI Projects May Be Canceled By 2027," Yahoo News Malaysia, 2026. [Online]. Available: https://malaysia.news.yahoo.com/why-40-agentic-ai-projects-131500758.html

[50] Zycus, "Agent Washing in Procurement AI: The '50+ Agents' Myth," 2026. [Online]. Available: https://www.zycus.com/blog/agentic-ai/agent-washing-procurement-ai

[51] Forrester, "The State Of Agentic AI In 2026: Companies Are Chasing, Few Are Catching," 2026. [Online]. Available: https://www.forrester.com/blogs/the-state-of-agentic-ai-in-2026-companies-are-chasing-few-are-catching/

[52] J. Ding and M. Liedtke, "Waymos blocked roads and caused chaos during San Francisco power outage," *AP News*, Dec. 21, 2025. Available: https://apnews.com/article/waymo-cars-san-francisco-power-outage-traffic-81e6a00aa2be6b804fe0bdfbcf07401f .

[53] "The Huge Problem Waymo Didn't See Coming," *The Atlantic*, Dec. 21, 2025. Available: https://www.theatlantic.com/technology/2025/12/waymo-robotaxi-san-francisco-blackout/685393/.